\documentclass{article}
\usepackage{ijcai26}

\usepackage{times}
\usepackage[utf8]{inputenc}
\usepackage[hidelinks]{hyperref}
\usepackage{url}
\usepackage{amsmath,amssymb}
\usepackage{amsthm}
\usepackage[small]{caption}   
\usepackage{subcaption}
\usepackage{graphicx}
\graphicspath{{arxiv26/}{./}}
\usepackage{booktabs}
\usepackage{siunitx}
\usepackage{float}

\usepackage{algorithm}
\usepackage{algpseudocode}

\usepackage{tcolorbox}
\tcbuselibrary{breakable}
\usepackage{fancyvrb}

\usepackage[switch]{lineno}

\usepackage{xspace}

\newcommand{\util}{\textbf{LISTEN-U}\xspace}
\newcommand{\tournament}{\textbf{LISTEN-T}\xspace}

\title{LISTEN to Your Preferences: \\
An LLM Framework for Multi-Objective Selection\setcounter{footnote}{1}\thanks{Accepted at IJCAI-ECAI 2026, the 35th International Joint Conference on Artificial Intelligence.}}

\author{
\mbox{Adam S. Jovine}$^{1\ast}$\and
\mbox{Tinghan Ye}$^{2\ast}$\and
\mbox{Francis Bahk}$^1$\and
\mbox{Jingjing Wang}$^1$\\
\mbox{Matthew Ford}$^1$\and
\mbox{David B. Shmoys}$^1$\and
\mbox{Peter I. Frazier}$^1$\\
\affiliations
$^1$Cornell University\\
$^2$Georgia Institute of Technology\\
\emails
asj53@cornell.edu,
joe.ye@gatech.edu,
\{feb47,jw2446,mtf62,dbs10,pf98\}@cornell.edu
}

\begin{document}

\maketitle

{\let\thefootnote\relax\footnotetext{$^\ast$\,Equal contribution.}}

\begin{abstract}
Human experts often struggle to select the best option from a large set of items with multiple competing objectives, a process bottlenecked by the difficulty of formalizing complex, implicit preferences. To address this, we introduce \textbf{LISTEN} (\textbf{L}LM-based \textbf{I}terative \textbf{S}election with \textbf{T}rade-off \textbf{E}valuation from \textbf{N}atural-language), an agentic LLM-based framework that treats the LLM as a decision-making agent capable of iteratively refining its internal preference model and taking actions (e.g., proposing utilities or selecting candidates) to maximize alignment with a user's implicit goals. To operate within LLM constraints like context windows and inference costs, we propose two iterative algorithms: \textbf{LISTEN-U}, which uses the LLM to refine a parametric utility function, and \textbf{LISTEN-T}, a non-parametric method that performs tournament-style selections over small batches of solutions. Evaluated on diverse tasks including flight booking, shopping, and exam scheduling, our results show LISTEN-U excels when preferences are parametrically aligned (a property we measure with a novel concordance metric), while LISTEN-T offers more robust performance overall. This work explores a promising direction for steering complex multi-objective decisions directly with natural language, reducing the cognitive burden of traditional preference elicitation. Code is available at \url{https://github.com/AdamJovine/LISTEN}; data is available at \url{https://huggingface.co/datasets/AdamJovine/LISTEN-benchmark}.
\end{abstract}

%
%

\section{Introduction}

We consider a human decision maker choosing among a large set of available items, such as airline flight itineraries or final exam schedules. The decision maker is aided by attributes describing the items (e.g., number of connections, price, total duration, departure time). Unfortunately, the decision maker lacks a precisely defined utility function over the attributes that would support the automatic selection of the best item. Instead, the decision maker must manually inspect many items to select the one they prefer most. This can lead to decision fatigue and poor decisions~\cite{schwartz2015paradox}.

This problem is common. Surveys across engineering and science domains~\cite{Farzane_2022,sharma2022comprehensive,gunantara2018review} argue that most real-world engineering design problems have multiple competing objectives. Individuals also face many options in day-to-day life, e.g., in online shopping, where combinatorial bundles of interacting products (flight legs arranged into itineraries, computer components arranged into a server) can explode in number.

Traditional solutions are often inadequate. Manual comparison of all options is time-consuming and prone to errors. Pareto plots displaying one attribute against another do not scale beyond a few attributes.
Multi-objective optimization algorithms (see, e.g., \cite{knowles2006parego,branke2008multiobjective})
allow screening solutions off the Pareto frontier; however, when there are many attributes, the number of Pareto-optimal solutions can be large.
Utility elicitation, preference learning, and other algorithmic search methods that require the decision-maker to express pairwise preferences between solutions
\cite{ObayashiJeongChiba2007,WangJinSchmittOlhofer2022,huber2025bayesian} or human-directed faceted search~\cite{Ozaki2024MultiObjectiveBayesian} can help, but still require significant human effort. The core difficulty is that {\it humans lack a time-efficient way to accurately articulate their preferences}.

Large language models (LLMs) open a new path here. With their ability to interpret text, LLMs enable zero-shot preference modeling, where a decision-maker's goals can be understood directly from a verbal description without expressing pairwise preferences. While recent research has begun integrating LLMs into preference learning, they are typically used as components within larger systems---to guide questioning~\cite{Lawless2023IWantItThatWay,Austin2024PEBOL}, extract preferences from reviews~\cite{Bang2025PURE}, or simulate user behavior~\cite{Okeukwu2025Community,Zhang2025UR4Rec}. However, how to best use LLMs to accelerate human selections over a large collection of items remains understudied.

\subsection{Contributions}
To address this gap, we introduce \textbf{LISTEN}: \textbf{L}LM-based \textbf{I}terative \textbf{S}election with \textbf{T}rade-off \textbf{E}valuation from \textbf{N}atural-language, an agentic framework in which an LLM acts as a decision-making agent that iteratively observes candidate solutions, reasons about trade-offs, updates an internal preference representation, and takes actions to refine the candidate set.

In our framework, a human expert first describes their priorities in natural language. Then, using iterative refinement, the algorithm selects its estimate of the best item.
Our approach must contend with limits on the LLM's context window and its ability to reason directly over large item lists, which can prevent the LLM from selecting the best item in a single call. Our approach must also limit the number of calls to the LLM to save computational and financial costs.

We introduce two algorithms in the LISTEN framework: \util, which makes decisions according to parametric utility functions adaptively generated by the LLM; and \tournament, which uses the LLM to compare solutions in a tournament-style selection mechanism.

We find that both methods perform at the level of the baselines or better. Our baselines span uniform random selection, an equal-weighted average over numerical attributes, single-call LLM ranking of all candidates, and human re-ranking.
\util significantly outperforms other methods when the held-out human rankings are in concordance with the parametric assumption of the utility function, as measured by a novel concordance metric developed in this work. When the human rankings are not in concordance, this method may underperform unless concordance is improved by adjusting the prompt or the utility form.
\tournament is not bound by parametric utility assumptions and provides robust performance across a range of problems.

\subsection{Related Work}

Our work contributes to the rapidly expanding application of LLMs in optimization, yet we address a distinct and often overlooked challenge. Much of the existing research focuses on the \textit{pre-solving phase}, from automatically formulating problems from natural language~\cite{ramamonjison2023nl4opt,ahmaditeshnizi2023optimus,xiao2023chain,huang2025orlm}, to configuring solver algorithms~\cite{lawless2025llms} and even positioning the LLM as a direct, formulation-free optimizer~\cite{yang2023large}. Complementing these efforts, other work evaluates the foundational knowledge of LLMs on core principles, such as primal-dual theory, which is crucial for reliable modeling and education~\cite{klamkin2025dualschool}. In contrast to these approaches, our work addresses the \textit{post-solution} challenge by providing a method for users to efficiently filter and navigate the large set of trade-off solutions from a multi-objective optimization, thus complementing the existing pipeline.

Our work is also related to research on how LLMs express preferences in pairwise comparisons, a concept foundational to Reinforcement Learning from Human Feedback (RLHF)~\cite{ouyang2022training} and the ``LLM-as-a-judge'' paradigm~\cite{zheng2023judging}. \cite{liu2024aligning} demonstrated that models differ substantially in their logical preference consistency, across dimensions such as transitivity, commutativity, and negation invariance, and that improving this consistency through methods such as their REPAIR method, leads to more stable, reliable, and higher-performing systems in downstream ranking and evaluation tasks. At the same time, more recent work warns us of potential pitfalls when treating LLMs as evaluators. \cite{gao2025scylla} shows some common pitfalls when using LLMs to simulate or replicate human behavior. Their results suggest that LLMs’ internal preference structures and responses are sensitive to extraneous artifacts (such as prompt framing, role assignment, or safety filters) and diverge from human behavior in unpredictable ways. We extend these results to new multi-objective settings, investigating whether these preference instabilities are amplified when navigating the complex trade-offs between competing objectives.

\begin{figure*}[!ht]
    \centering
    \includegraphics[width=0.6\linewidth]{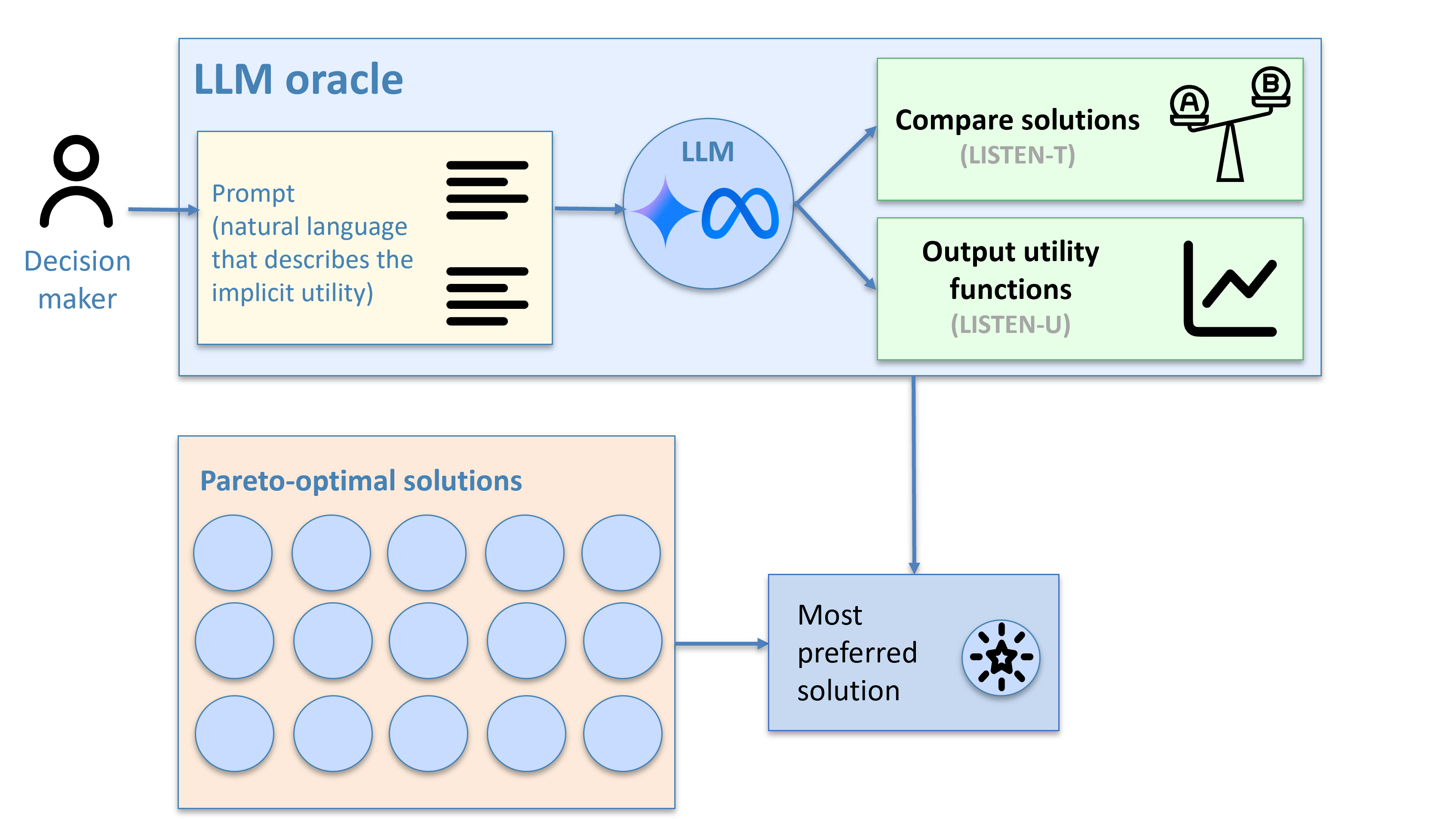}
    \caption{LISTEN as an agentic decision loop: an LLM agent iteratively refines preferences or selects champions to identify the most preferred solution.
    }
    \label{fig:schematic}
\end{figure*}

\section{Problem Description}
\label{sec:problem}

We consider the problem of selecting a single preferred item from a large collection of candidates. In many applications, this collection represents the set of Pareto-optimal solutions generated by a multi-objective optimization algorithm. Formally, let $\mathcal{S} = \{s_1, s_2, \dots, s_N\}$ be a set of $N$ items. Each item $s_i \in \mathcal{S}$ is described by a tuple of $d$ attributes, $s_i = (a_{i1}, \dots, a_{id})$. The attributes can be of mixed types; each attribute $a_{ij}$ belongs to a domain $\mathcal{D}_j$ that can be \textbf{numerical} (e.g., $\mathbb{R}$ for price), \textbf{categorical} (e.g., a set of airline names), or \textbf{textual} (e.g., a free-form product description). We are also given a natural language utterance, $U$, that articulates a human decision-maker's preferences over these attributes.

Our goal is to identify the item $s^\star \in \mathcal{S}$ that best satisfies the preferences described in $U$. We assume access to a pre-trained LLM, which acts as a preference-driven decision agent, maintaining an evolving internal preference state and producing actions (utility updates or selections) conditioned on its observation history. The core challenge is to design an algorithm that finds $s^\star$ while adhering to a limited budget of calls to the LLM. This setting is motivated by scenarios where $N$ and/or $d$ are large, making it unrealistic for a human to exhaustively inspect all items and unlikely for an LLM to process the entire set in a single context window. We treat $U$ as one decision-maker's articulation of their own preferences, and target alignment with that individual user.

\section{Methodology}

To solve the selection problem defined in Section~\ref{sec:problem}, we introduce \textbf{LISTEN} (\textbf{L}LM-based \textbf{I}terative \textbf{S}election with \textbf{T}rade-off \textbf{E}valuation from \textbf{N}atural-language). LISTEN is an agentic, iterative decision framework in which an LLM acts as a closed-loop preference agent.  As illustrated in Figure~\ref{fig:schematic}, at each step, the agent observes a subset of candidate solutions, reasons about trade-offs using the user’s natural-language priorities, updates an internal preference representation, and takes an action, either proposing a refined utility function (LISTEN-U) or selecting a batch winner (LISTEN-T), to guide the next iteration.
This iterative design is crucial for navigating large solution spaces while respecting the LLM's inherent context window and query budget limitations.

\subsection{Prompting the Preference Oracle}
\label{ssec:prompt_structure}

The core of our framework is using an LLM to simulate the decision-maker's preferences in a zero-shot manner. We do not fine-tune the LLM; instead, all queries follow a structured prompt that encapsulates the decision context. Each prompt sent to the LLM consists of five key components:
\begin{enumerate}
    \item \textbf{Persona Context}: Assigns a role to the LLM (e.g., ``You are a University registrar'').
    \item \textbf{Attribute Definitions}: Defines the attributes of the items (e.g., ``Conflicts: students with two exams at the same time'').
    \item \textbf{User Priorities}: The natural language utterance $U$ describing the user's goals (e.g., ``prioritize minimizing back-to-back exams'').
    \item \textbf{Solutions}: The candidate item(s) to be evaluated in the current step.
    \item \textbf{Format Instructions}: Specifies the desired output format (e.g., a JSON object).
\end{enumerate}
This consistent structure provides the LLM with all the necessary context to act as the preference oracle in any given algorithmic step, whether it is refining a utility function or selecting a champion from a batch.

From an agentic perspective, each prompt instantiates the agent’s current belief state (encoded via weights or champions), while the LLM’s output corresponds to an action that updates this state or advances the selection policy. This explicit observe–reason–act loop situates LISTEN as a lightweight, tool-free decision agent.

\subsection{LISTEN-U: Iterative Utility Refinement}
\label{ssec:listen_u}

Our first algorithm, LISTEN-Utility (\util), is a parametric approach. It assumes a linear utility function, $u(s) = \mathbf{w}^\top s^{\text{num}}$, defined over the encoded feature vector $s^{\text{num}}$: genuinely numerical fields (e.g., price, battery life) and, when an integer-to-category mapping is supplied to the LLM in context, label-coded categorical fields. Free-text descriptions and other non-scored attributes inform the LLM's reasoning but are not part of the scoring function. The agent iteratively refines its internal utility representation $\mathbf{w}$ to identify the best item, which serves as a compact belief state over the user’s latent preferences, as detailed in Algorithm~\ref{alg:listen-u}.

\begin{algorithm}[!t]
\caption{LISTEN-U: Iterative Utility Refinement}
\label{alg:listen-u}
\begin{algorithmic}[1]
\Require Solution set $\mathcal{S}$, max iterations $T$, LLM oracle $\mathcal{L}$, preference utterance $U$
\State \textbf{Initialization:}
\State Construct initial prompt $p_1$ with $U$ and definitions of all attributes
\State Elicit initial weight vector $\mathbf{w}_1$ for numerical attributes from $\mathcal{L}(p_1)$
\State For each $s_i \in \mathcal{S}$, let $s_i^{\text{num}}$ be the vector of its numerical attributes
\State Let $\tilde{s}_i^{\text{num}}$ be the version of $s_i^{\text{num}}$ with attributes normalized to $[0,1]$
\State $s^\star \gets \arg\max_{s_i \in \mathcal{S}} \mathbf{w}_1^\top \tilde{s}_i^{\text{num}}$

\State \textbf{Iterative Refinement:}
\For{$t \gets 2$ to $T$}
    \State Construct refinement prompt $p_t$ using $U$, $\mathbf{w}_{t-1}$, and all attributes (numerical and non-numerical) of the \textit{unnormalized} solution $s^\star$
    \State Elicit refined weight vector $\mathbf{w}_t$ for the numerical attributes from $\mathcal{L}(p_t)$
    \State $s^\star \gets \arg\max_{s_i \in \mathcal{S}} \mathbf{w}_t^\top \tilde{s}_i^{\text{num}}$
\EndFor

\State \Return final weight vector $\mathbf{w}_T$ and solution $s^\star$
\end{algorithmic}
\end{algorithm}

The algorithm operates in two phases. In the \textbf{initialization phase} (lines 2-6), the LLM is prompted to define an initial weight vector, $\mathbf{w}_1$, for the numerical attributes based on the user’s utterance. To score the items, the numerical attributes of every solution in $\mathcal{S}$ are first normalized to a common $[0,1]$ scale. The algorithm then computes the utility of each item by taking the dot product of the weights and the normalized numerical attribute vector, then selecting the highest-scoring item as the initial best solution, $s^\star$.

Next, the \textbf{iterative refinement phase} (lines 8-12) begins. In each iteration, the prompt for the LLM is constructed using the current weight vector $\mathbf{w}_{t-1}$ alongside the complete, \textbf{unnormalized} description of the current best solution, $s^\star$, including both its numerical and non-numerical attributes. Presenting the true values and full context is crucial for the LLM to reason about real-world trade-offs (e.g., ``the price is too high \textit{for this particular brand}''). The LLM is asked to critique this solution and propose a refined weight vector, $\mathbf{w}_t$. Once the LLM returns the updated weights, the algorithm again scores the entire solution set using the consistently \textbf{normalized} numerical attributes to select the new best solution.

\subsection{LISTEN-T: Tournament-Based Selection}
\label{ssec:listen_t}

Our second algorithm, LISTEN-Tournament (\tournament), is a non-parametric method that emulates a tournament to efficiently search the solution space. It is designed for a budget of $T \geq 3$ calls to the LLM, as detailed in Algorithm~\ref{alg:listen-t}. In agentic terms, LISTEN-T implements a selection policy in which the agent adaptively samples, evaluates, and promotes candidate actions (solutions) through a stochastic elimination process.

\begin{algorithm}[!t]
\caption{LISTEN-T: Tournament-Based Selection}
\label{alg:listen-t}
\begin{algorithmic}[1]
\Require Solution set $\mathcal{S}$, batch size $B$, max iterations $T \geq 3$, LLM oracle $\mathcal{L}$, utterance $U$

\State \textbf{Preliminary Rounds:}
\State $R \gets T - 1$ \Comment{Number of preliminary rounds}
\State $\mathcal{K} \gets \emptyset$ \Comment{Initialize set of batch champions}
\For{$j \gets 1$ to $R$}
    \State $\mathcal{C}_j \gets \text{Sample}(\mathcal{S}, B)$ \Comment{Sample a batch of size B}
    \State $c_j \gets \text{LLM-Choose}(\mathcal{C}_j, U)$ \Comment{LLM selects batch champion}
    \State $\mathcal{K} \gets \mathcal{K} \cup \{c_j\}$
\EndFor

\State \textbf{Final Playoff:}
\State $s^\star \gets \text{LLM-Choose}(\mathcal{K}, U)$ \Comment{Select winner from champions}
\State \Return solution $s^\star$
\end{algorithmic}
\end{algorithm}

The algorithm proceeds in two stages using a total of $T$ calls to the LLM. First, in the \textbf{preliminary rounds} (lines 2-8), the algorithm conducts $R = T-1$ rounds. In each round, it samples a batch of $B$ items uniformly at random without replacement from $\mathcal{S}$ and prompts the LLM to select the single most preferred item from that batch. This winning item, or ``batch champion,'' is added to a set of champions, $\mathcal{K}$.

Second, in the \textbf{final playoff} (line 10), the LLM is presented with the set of all $R$ champions collected from the preliminary rounds and is asked to select the single overall winner. By requiring $T \geq 3$, this structure ensures the final playoff is a meaningful comparison between at least two distinct batch champions, allowing the LLM to make a final, decisive trade-off.

\section{Experiments}
\label{sec:experiments}

We conduct a series of experiments to evaluate the effectiveness of our LISTEN framework in identifying a user's most preferred item from a large set of multi-attribute options. We ground our evaluation in three realistic decision-making domains: (i) selecting a flight itinerary from a commercial flight search engine, (ii) choosing a product from an e-commerce website, and (iii) scheduling university final exams from a set of Pareto-optimal solutions generated by an optimization solver.

In the following sections, we detail our methodology for collecting real human preference data (Section~\ref{sec:data-collection}), describe the experimental setup (Section~\ref{sec:setup}), outline the baseline algorithms for comparison (Section~\ref{sec:baselines}), and define our evaluation metrics (Section~\ref{sec:metrics}), before presenting the main results in Section~\ref{sec:results}.

\subsection{Human Preference Data Collection}
\label{sec:data-collection}

To establish a ground truth for evaluating our algorithms, we created a human preference dataset for each of our three domains. The data was generated by a decision-maker (DM) with significant experience in each respective task, following a two-step protocol. First, the DM articulated their decision-making criteria in natural language, framed as a detailed directive to an assistant. This text serves as the preference utterance, $U$, for each problem. Second, the DM meticulously ranked a subset of items from the full candidate set according to these same preferences. This process yields a dataset comprising a high-level natural language goal ($U$) paired with a corresponding ground-truth ranked list of items. This ranking is designed to capture the nuanced, implicit trade-offs an expert would make in a real-world scenario, providing a realistic benchmark for evaluating an algorithm's alignment with complex human preferences. The specifics of each dataset are detailed below.

\paragraph{Flights Dataset.}
We generated a dataset of realistic flight itineraries using the \texttt{fast\_flights} Python package. For a given origin, destination, and travel date, the package returns a collection of available flights. Each itinerary is described by primary attributes such as airline, departure/arrival times, duration, number of layovers, and price. We augmented these with engineered features, such as the ground travel distance from the airport to a user's specified final location. For the flight datasets, categorical attributes such as airline were one-hot encoded for LISTEN-U and numerical baselines; for all LLM prompts, these attributes were retained as their original text to provide full context.

We designed two distinct experimental scenarios to test a range of preferences: \textbf{Flights CHI $\rightarrow$ NYC}: A student flying from Chicago, IL to New York, NY, and \textbf{Flights Ithaca $\rightarrow$ Reston}: A professor flying from Ithaca, NY to Reston, VA.

For each scenario, we crafted a unique preference utterance ($U$) written in conversational English to emulate a real-world directive. For instance, a preference might be articulated as: \textit{``If I fly in on Friday, I prefer to fly in early enough to get a good night of sleep... I prefer a direct flight. I have no preference for airline.''} The full set of preference utterances is provided in the extended arXiv version.

\paragraph{Headphones Dataset.}
Our second domain involves an e-commerce scenario where a college student is shopping for headphones. The preference utterance ($U$) for this task specifies a desire for a reliable pair of daily-use headphones that are over-ear, wireless, have active noise cancellation, a built-in microphone, and long battery life.

To generate the dataset, we collected public product information for a range of headphones from Amazon.com using the \texttt{Scrapingdog} API. Each product is an item in our set $\mathcal{S}$, characterized by a mix of attribute types. \textbf{Numerical} attributes include price, average rating, review count, battery life (hours), and weight. \textbf{Categorical} attributes include the headphone type (e.g., ``Over-ear''), connectivity, noise cancellation mode, water resistance, and the presence of a microphone. Additionally, each item contains \textbf{textual} fields such as the product name, brand, and full description, which provide rich context for the LLM's evaluation. These categorical fields are label-encoded in the released feature table, with qualitative mappings supplied in the prompt header.

\paragraph{Final Exam Scheduling Dataset.}
Our third domain addresses a large-scale university final exam scheduling problem. We generated a set $\mathcal{S}$ of 4,938 diverse, Pareto-optimal schedules using the ParEGO multi-objective optimization algorithm~\cite{knowles2006parego}, based on the mixed-integer programming methods of~\cite{JoeYeCornell}. The quality of each schedule is evaluated against a comprehensive set of attributes to be minimized, listed in the extended arXiv version. The primary attribute is \textbf{Direct Conflicts}, counting students with overlapping exams. A suite of attributes addresses \textbf{Exam Overload} by penalizing high-density patterns, such as five (\textit{Quints}), four (\textit{Quads}), or three (\textit{Triples}) consecutive exams, as well as near-consecutive patterns (e.g., \textit{Four in Five Slots}). General student fatigue is measured by the total number of \textbf{Back-to-Back} exams, which are separated into \textit{Evening-to-Morning} and all \textit{Other} cases. Finally, the \textbf{Schedule Spread} is measured by the \textit{Average Time of Last Exam}. Together, these attributes capture the nuanced trade-offs between student workload, fairness, and logistical constraints.

\paragraph{A Dataset Diagnostic: Concordance.}
To analyze why our parametric (LISTEN-U) and non-parametric (LISTEN-T) algorithms perform differently across datasets, we introduce a metric to quantify the complexity of the underlying preference structures. This metric, computed for each dataset independently of any algorithm, characterizes the intrinsic ``difficulty" of reaching the human's top picks using a linear utility function. To compute it, we generate 10,000 random linear utility functions, $u(s) = \mathbf{w}^\top s^{\text{num}}$, by sampling each weight $w_i$ independently from $\mathcal{U}[-1, 1]$. The concordance score is the fraction of these random functions for which the optimal item, $\arg\max_s u(s)$, lies within the $m$-item human-ranked set (see $m$ in Table~\ref{tab:difficulty}). This measures how broadly the human's top picks are reachable by linear utilities, not just whether the single \#1 pick is. A low concordance value, as seen for the exam scheduling dataset in Table~\ref{tab:difficulty}, suggests that the human preference is complex and not easily selected by a linear model. A high concordance value, as in Flights Ithaca $\rightarrow$ Reston, suggests a much more linearly separable problem. Table~\ref{tab:difficulty} shows that our datasets span a wide range of concordance values, indicating varying difficulty for linear preference models. This provides a robust testbed for comparing our parametric (LISTEN-U) and non-parametric (LISTEN-T) algorithms.

\begin{table*}[!ht]
\centering
\sisetup{separate-uncertainty=true}
\begin{tabular}{l S[table-format=1.4(4)] ccc}
\toprule
\textbf{Dataset} & {\textbf{Concordance}} & \textbf{Total Items ($N$)} & \textbf{Ranked Items ($m$)} & \textbf{Ranked Prop. ($m/N$)} \\
\midrule
Exam Scheduling & 0.0015 \pm 0.0008 & 4938 & 11 & 0.002 \\
Flights CHI $\rightarrow$ NYC
       & 0.0025 \pm 0.0011 & 903  & 16  & 0.018 \\
Headphones      & 0.0552 \pm 0.0047 & 77   & 15  & 0.195 \\
Flights Ithaca $\rightarrow$ Reston
      & 0.3357 \pm 0.0095 & 216  & 12  & 0.056 \\
\bottomrule
\end{tabular}
\caption{Dataset statistics and the Concordance metric. Concordance measures the alignment of human preferences with random linear utilities (a lower value indicates a more difficult problem for linear methods). Values for the Concordance column are reported as mean $\pm$ 2SE (approximating a 95\% CI).}
\label{tab:difficulty}
\end{table*}

\subsection{Experimental Setup}
\label{sec:setup}

Across all experiments, we use two LLMs as our preference oracle: \texttt{Llama-3.3-70B-Versatile}~\cite{llama3_2024} and \texttt{Gemini 2.5 Flash-Lite}~\cite{gemini2_5_2025}. To ensure the robustness of our results, all algorithmic runs are replicated 40 times with different random seeds for generation, while keeping the prompt consistent for each replication. Code, prompts, and the human-preference datasets are available at \url{https://github.com/AdamJovine/LISTEN-IJCAI}.

The ground truth for evaluation is derived from human expert rankings. For each dataset, it is impractical to rank the entire set of $N$ items, so an expert manually ranked a subset of the most relevant items ($m \ll N$). This ranked list serves as the ground truth for measuring how well an algorithm's selected item aligns with nuanced human preferences.

\subsection{Baseline Methods}
\label{sec:baselines}

We compare the performance of our LISTEN algorithms against the following four baseline methods.

\paragraph{Uniform Random Selection (\texttt{baseline/random}).}
This non-informative baseline selects an item uniformly at random from the full candidate set~$\mathcal{S}$. It is used to establish a lower bound on performance and quantify the improvement gained by any guided selection strategy.

\paragraph{Normalized Average Score (\texttt{baseline/zscore-avg}).}
This method considers only the numerical attributes of the items. For each numerical attribute, it first standardizes the values across the entire set $\mathcal{S}$ to have zero mean and unit variance (i.e., calculates the z-score). The scores for attributes that are to be minimized (e.g., price) are then negated. The algorithm selects the item with the highest average standardized score across all numerical attributes. This baseline represents a simple, non-learned linear utility function that weights all normalized metrics equally and ignores all categorical or textual information.

\paragraph{Full Batch
(\texttt{baseline/full$\_$batch}).}
This method works like a single preliminary iteration of LISTEN-T, with the entire dataset passed into the prompt in this single iteration. The LLM is asked to rank the items. We randomize the seed used for sampling in the LLM and the order the items are presented in each replication of the method. The top-ranked item from each replication is reported.

\paragraph{Human Rankers
(\texttt{baseline/human$\_$rerank}).}
To compare against a human baseline with access to textual attributes in addition to numerical, we tasked human rankers different from the decision-maker creating the ground truth labels with assigning rankings. These humans were given a description of the attributes of items in each dataset, the decision-maker's preference utterance, and a system allowing them to filter and sort by multiple attributes at a time. We tasked them with ranking the top 20 items they believed best matched the preference utterance, while ensuring they explored many high-quality items. However, like the other baselines and methods, we only report their final top-ranked item.

\subsection{Evaluation Metric: Normalized Average Rank}
\label{sec:metrics}

We evaluate algorithm performance primarily using the rank of the selected item within the human-generated ground-truth ranking.
For a given dataset with $N$ total items, let the ground-truth list contain $m$ items ranked by a human expert ($1, 2, \dots, m$). For an item $s^\star$ selected by an algorithm, its rank is determined as follows: If $s^\star$ is within this top-$m$ list, its rank is its position in that list. If $s^\star$ is not in the list, it is considered unranked. All $N-m$ unranked items are assigned a shared average rank of $(m+1+N)/2$, representing the mean of the available ranks from $m+1$ to $N$. To ensure a consistent scale across datasets of different sizes, we normalize this rank by dividing by the total number of items, $N$. The final metric is a value in $(0, 1]$, where a lower score indicates better performance.

\subsection{Results \& Discussions}
\label{sec:results}

\begin{figure*}[!ht]
    \centering
    \begin{subfigure}{0.45\linewidth}
    \centering
    \includegraphics[width=\linewidth]{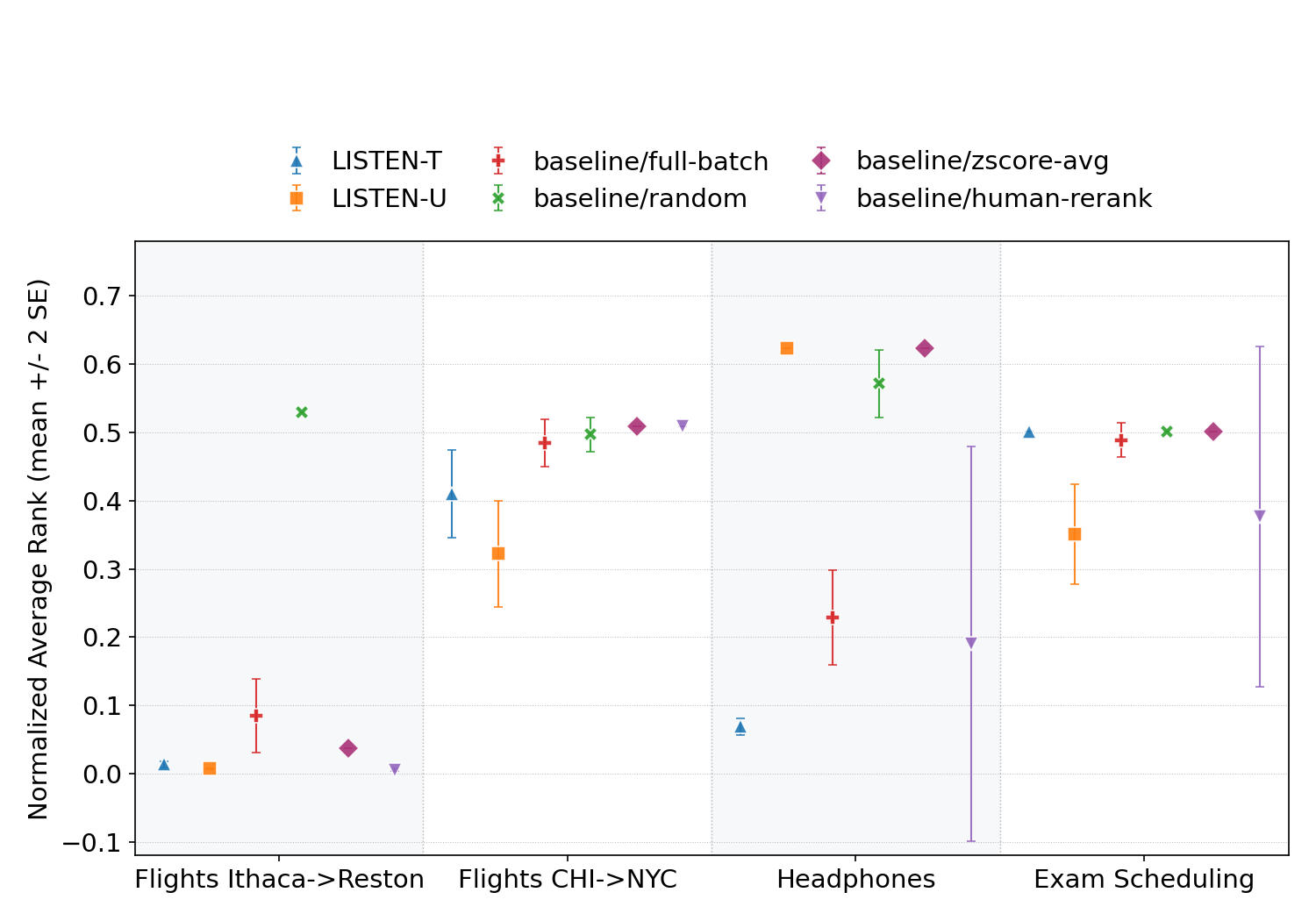}
    \caption{LLaMA}
    \label{fig:llama}
    \end{subfigure}
    \hfill
    \begin{subfigure}{0.45\linewidth}
    \centering
    \includegraphics[width=\linewidth]{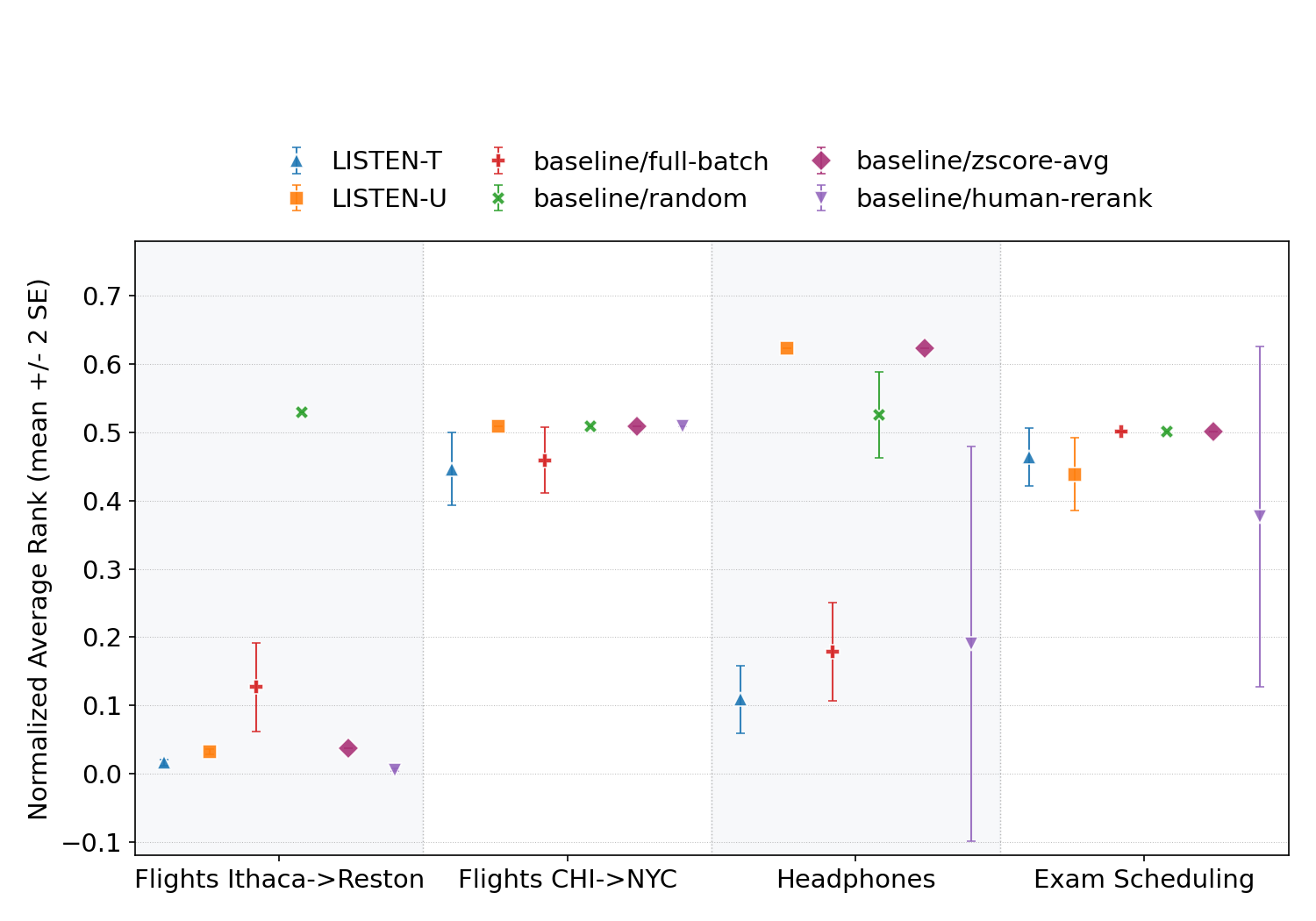}
    \caption{Gemini}
    \label{fig:gemini}
    \end{subfigure}
    \caption{Performance of LISTEN algorithms and baselines on four datasets, showing the Normalized Average Rank (lower is better) of the top ranked item by each method. The LISTEN methods used 25 iterations and batch size 32. All algorithms have $n=40$ runs except baseline/human-rerank ($n=4$) and baseline/zscore-avg ($n=1$).}
    \label{fig:unified}
\end{figure*}

We present experimental evaluation of the LISTEN framework in Figure~\ref{fig:unified}. We then analyze the comparative performance of our algorithms and conclude with ablation studies.

\subsubsection{Comparisons to Baselines}

The LISTEN methods tend to outperform the non-human baselines. We now focus on the full-batch and human-rerank baselines, as they are the most competitive and also give insights into the complexity of our problem setting.

\paragraph{\texttt{baseline/full$\_$batch:}}
The LISTEN methodology offers an improvement in ranking performance relative to the naive approach of asking the LLM to rank all items at once, as demonstrated in Figure~\ref{fig:unified}. The smaller ranking tasks performed in each iteration of LISTEN approaches reduce the complexity of the task. In addition, running multiple iterations where the most promising items are compared against each other gives the LLM another chance to focus on the tradeoffs present in high-quality options and make a refined top choice. We also note that for some datasets and LLM combinations, full-batch ranking would be practically infeasible due to not fitting in the LLM's context window.

\paragraph{\texttt{baseline/human$\_$rerank:}}
In comparison with the human-rerank baseline, we must first acknowledge the large variability in its performance. One reason for this is that humans had a higher dispersion of their top-choice being ranked vs unranked compared to LISTEN methods. Recall from Section~\ref{sec:metrics} that unranked items receive a normalized average rank of $\approx 0.5$. The best human ranks tended to be very strong; however, many humans' top choices were not in the ranked set from the decision-makers' ground truth rankings. Another reason is we only had 4 samples for humans. This is the reason for the large variance in normalized average ranks.

However, this highlights the ambiguity in the distribution of preferences that humans associate with a prompt. We see that LISTEN methods more consistently chose items that were ranked than humans, but perhaps may be less likely to optimize the very fine tradeoffs the decision-maker prefers.

\subsubsection{Validating the Concordance Metric}

\begin{figure}[!ht]
    \centering
    \includegraphics[width=0.75\linewidth]{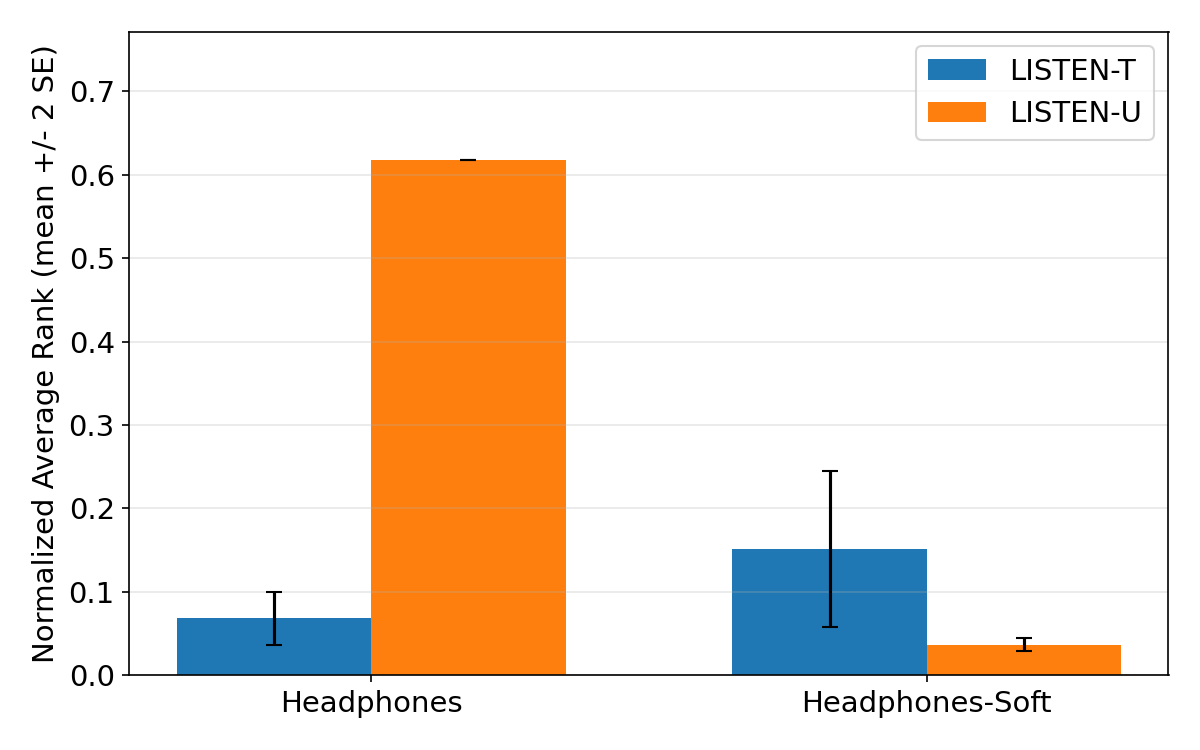}
    \caption{Performance of LISTEN-U and LISTEN-T on the default \texttt{Headphones} prompt and a softer variant \texttt{Headphones-Soft} that drops the hard constraints, which has higher concordance. All algorithms have $n=40$ runs.}
    \label{fig:high_vs_low_concordance}
\end{figure}

To directly test the relationship between preference complexity and our concordance metric, we conducted a targeted experiment on the Headphones dataset. The default \texttt{Headphones} prompt (Table~\ref{tab:difficulty}) contains hard constraints such as a strict budget and required categorical features. We created a softer variant, \texttt{Headphones-Soft}, by dropping these constraints. Removing these threshold-based requirements makes the preference more linearly representable, causing the Concordance score to rise from \textbf{0.055} to \textbf{0.244}. This rise is expected, as linear utility functions struggle to model concrete threshold-based requirements.

As predicted by this change in concordance, Figure~\ref{fig:high_vs_low_concordance} shows a significant increase in performance for LISTEN-U on the \texttt{Headphones-Soft} version. In contrast, the non-parametric LISTEN-T maintained its robust performance across both prompts. This result validates that our concordance metric is an effective predictor for the performance of a parametric approach like LISTEN-U.

\subsection{Observed Robustness to LLM Choice and Batch Size}

\begin{figure}[!ht]
    \centering
    \begin{subfigure}{0.88\linewidth}
        \centering
        \includegraphics[width=\linewidth]{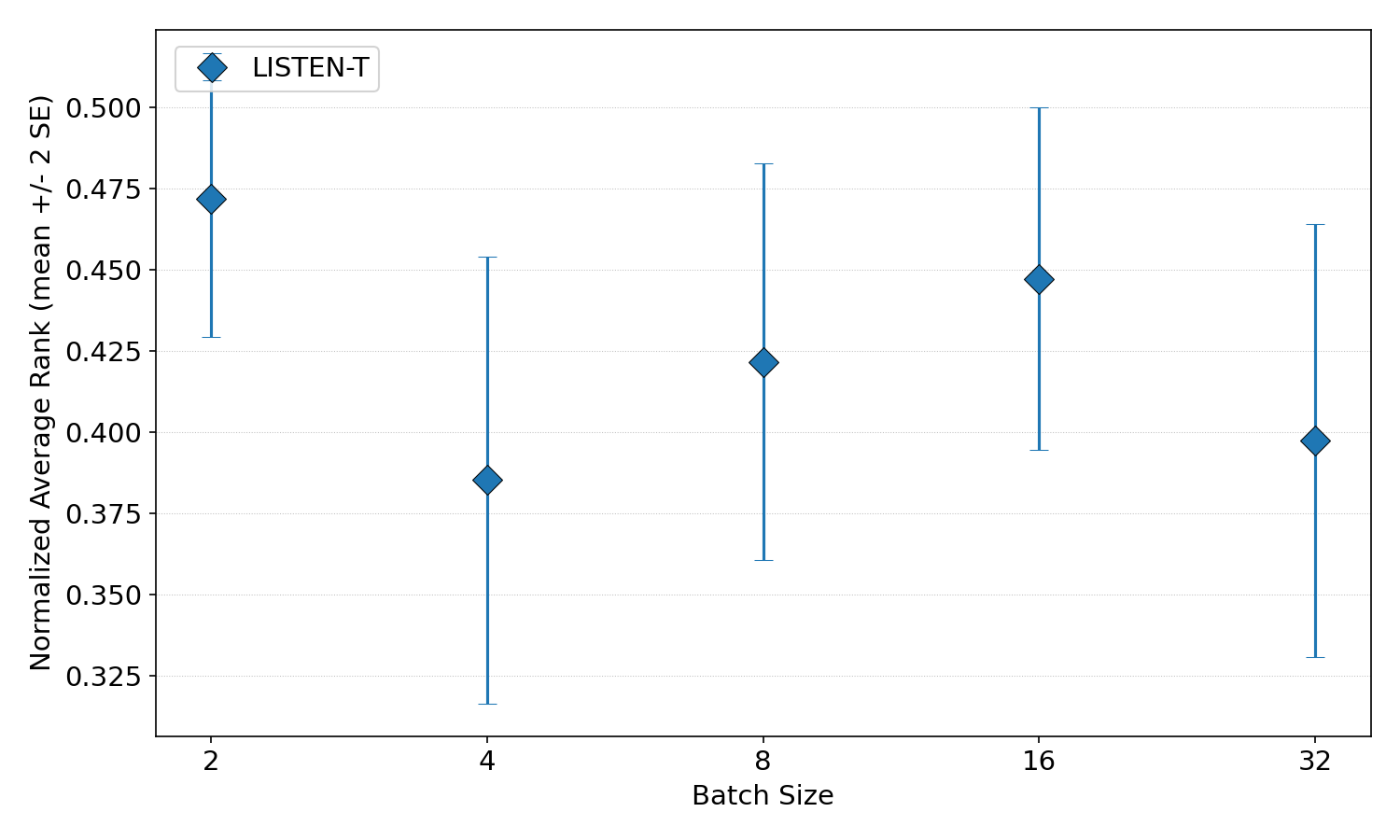}
        \caption{LLaMA, Flights CHI $\rightarrow$ NYC}
    \end{subfigure}

    \begin{subfigure}{0.88\linewidth}
        \centering
        \includegraphics[width=\linewidth]{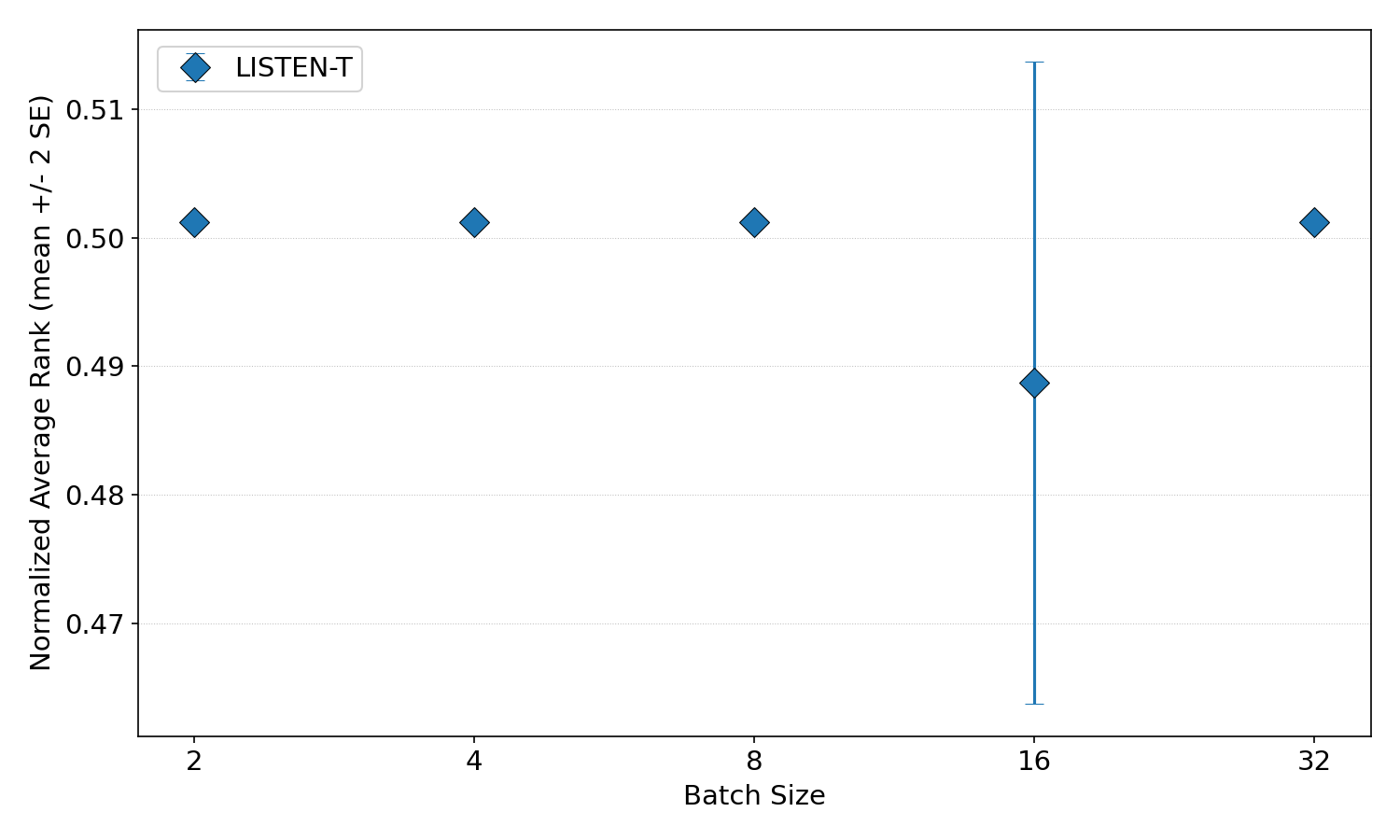}
        \caption{LLaMA, Exam Scheduling}
    \end{subfigure}
    \caption{Batch size effect on LISTEN-T ($n=40$).}
    \label{fig:bat}
\end{figure}

We observe a surprisingly high level of consistency in performance across batch sizes for each (LLM, dataset) pair in our experiments, as seen in Figure~\ref{fig:bat}. Our full experimental results across all batch sizes (LLaMA and Gemini, batch sizes 2--32) are included in the extended arXiv version. There would be value in further analysis to see how robust this finding is across even larger batch sizes.

\section{Conclusion}

In this work, we introduced \textbf{LISTEN}, an agentic decision-support framework that uses large language models as preference-driven agents for multi-objective selection. By maintaining an evolving internal preference representation and acting iteratively on candidate sets, LISTEN turns natural-language goals into concrete decisions. Our experiments show that LLMs can successfully translate high-level natural language goals into high-quality selections from large, complex sets of items. We also introduced a concordance metric that predicts when the parametric approach is likely to succeed or fail based on the problem's preference structure.

That said, our study has several limitations that suggest clear avenues for future work.

\paragraph{Complexity of Human Rankings.}
The ground-truth rankings used in our evaluation reflect the preferences of a single expert for each domain and may not generalize broadly. We also noticed the high variability of our human re-rankers with respect to the ground-truth rankings. Future work could gather more data to understand the variability in preferences corresponding to an utterance and return single items or sets of items that satisfy the diverse potential preferences.

\paragraph{Generalization Across Domains.}
While we evaluated LISTEN on three distinct domains, its performance on a wider range of problems, such as apartment hunting, healthcare scheduling, or logistics planning, is yet to be tested at scale. Expanding to more varied and larger-scale datasets would further validate our findings.

\paragraph{Richer Utility Representations.}
The LISTEN-U algorithm's reliance on a linear utility function is a key limitation for problems with highly non-linear preferences. Future work should explore more expressive utility representations (e.g., non-linear or piecewise functions) and hybrid methods that combine LLM guidance with structured optimization.

\section*{Acknowledgements}
MF, FB, and PIF were supported by ONR 
N00014-22-1-2763 and N00014-25-1-2181. MF and PIF were supported in part by Moebius Solutions, Inc.

\bibliographystyle{named}
\bibliography{refs}

\onecolumn
\clearpage
\appendix
\renewcommand{\thesection}{Appendix~\Alph{section}}
\renewcommand{\thesubsection}{\Alph{section}.\arabic{subsection}}
\section{Attributes}
\label{sec:attributes}

This section details the attributes used across the three evaluation scenarios. The specific objectives for the final exam scheduling problem, the flight dataset, and the headphones dataset are documented in Table \ref{tab:exams}, Table \ref{tab:metrics}, and Table \ref{tab:headphones}, respectively.

\begin{table}[!ht]
\centering
\begin{tabular}{l p{0.6\columnwidth}}
\hline
\textbf{Attribute} & \textbf{Description} \\ \hline
Conflicts & Instances of a student having two or more exams in the same time slot. \\
Quints & Instances of a student having five exams in consecutive time slots. \\
Quads & Instances of a student having four exams in consecutive time slots. \\
Four in Five Slots & Instances of a student having four exams within five consecutive time slots. \\
Triple in 24h (no gaps) & Instances of a student having three back-to-back exams in a 24-hour period. \\
Triple in Same Day (no gaps) & Instances of a student having three back-to-back exams on the same day. \\
Triples & Triple in 24h (no gaps) + Triple in Same Day (no gaps).\\
Three in Four Slots & Instances of a student having three exams within four consecutive time slots. \\
Evening/Morning B2Bs & Instances of a student having an exam in the last slot of one day and the first slot of the next day. \\
Other B2Bs & All other instances of a student having exams in adjacent time slots. \\
Back-to-backs & Evening/Morning B2Bs + Other B2Bs.\\
Two in Three Slots & Instances of a student having two exams within three consecutive time slots.\\
Average Time of Last Exam & Average time slot in which students take their last exams. \\
\hline
\end{tabular}
\caption{Final Exam Scheduling Conflict Attributes}
\label{tab:exams}
\end{table}

\begin{table}[!ht]
\centering
\begin{tabular}{l p{0.6\columnwidth}}
\hline
\textbf{Attribute} & \textbf{Description} \\ \hline
Name & Name of airline operating the flight. \\
Origin & Origin airport. \\
Destination & Destination airport. \\
Departure Time & Time of departure from origin airport. \\
Arrival Time & Time of arrival at destination airport. \\
Duration & How long the flight is. \\
Stops & Number of layover stops.\\
Price & Cost of the flight. \\
dis\textunderscore from\textunderscore origin & Distance of origin airport from where the customer prefers (lower is better). \\
dis\textunderscore from\textunderscore dest & Distance of arrival airport from where the customer prfers (lower is better).\\
departure\textunderscore seconds & Time of departure since a fixed date in seconds.\\
arrival\textunderscore seconds & Time of arrival since a fixed date in seconds. \\
duration\textunderscore min & Duration of total flight in minutes (lower is better). \\
\hline
\end{tabular}
\caption{Flights Attributes}
\label{tab:metrics}
\end{table}

\begin{table}[!ht]
\centering
\begin{tabular}{l p{0.6\columnwidth}}
\hline
\textbf{Attribute} & \textbf{Description} \\ \hline
Product Name & Name of the headphone model. \\
Brand & Manufacturer. \\
Price & Cost in U.S. dollars. \\
Type & Headphone design (e.g., over-ear, in-ear, on-ear). \\
Connectivity & Connection type (wired vs.\ wireless). \\
Noise Cancellation & Noise reduction method (active vs.\ passive). \\
Battery Life & Battery duration, measured in hours. \\
Bluetooth Version & Bluetooth version, with higher values indicating newer technology. \\
Driver Size & Diameter of the audio driver, measured in millimeters. \\
Weight & Physical weight of the headphones, measured in ounces. \\
Water Resistance & Protection against dust and water, described qualitatively and via the IPXX rating system (higher values indicate stronger resistance). \\
Microphone & Presence of built-in microphone. \\
Review Rating & Average customer rating score. \\
Review Count & Total number of customer reviews. \\
Description & Marketing text describing the product. \\
\hline
\end{tabular}
\caption{Headphones Attributes}
\label{tab:headphones}
\end{table}

\section{Prompts} \label{app:prompts}
This section includes representative examples of the variation of prompts we utilized. As mentioned in Section 3.1, each prompt consists of five parts (persona context, attribute definitions, user priorities, solutions, and format instructions). Unless otherwise noted, every result reported in the main paper uses these parts in the canonical ordering Persona Context $\rightarrow$ User Priorities $\rightarrow$ Attribute Definitions, which is the ordering reproduced in the example prompts below; the prompt-order ablation in Section~\ref{sec:prompt-order} reports the other five permutations.
The first three parts are different in each setting and we give examples
below in figures (Figure \ref{fig:utility_prompt}, Figure \ref{fig:flight_prompt_hard}, \ref{fig:soft_headphones_prompt}, and Figure \ref{fig:strict_headphones_prompt}).
The fourth and fifth parts are relatively generic across settings. The fourth part (solutions) pulls the solution from the previous iteration, if any. The fifth part of the instructions, which covers output format, has different structures for LISTEN-U and LISTEN-T. For LISTEN-U, the task is to output the weights for numerical features in JSON format. For LISTEN-T, the task is to identify the single best solution from the options provided and return it using the exact format: \texttt{FINAL A (B, C ...)}.

\subsection{Exam Scheduling}
Only the first three parts of the prompt vary significantly, so Figure \ref{fig:utility_prompt} shows these parts of the prompt for the exam scheduling problem. The first line, ``You are an expert university registrar..." provides the persona context. The next section states the user priorities (``minimize student stress and administrative burden by optimizing a specific set of attributes..."). The attribute definitions are then listed in bulleted form below the priorities.

\begin{tcolorbox}[title=Exam Scheduling Prompt,
                  colback=gray!5!white, colframe=gray!70!black,
                  sharp corners, boxrule=0.6pt, width=\textwidth,
                  breakable]
\begin{verbatim}
You are an expert university registrar choosing the better final-exam schedule.

Policy guidance: To generate an optimal final examination schedule that
minimizes student stress and administrative burden by optimizing a specific
set of metrics according to a tiered priority system.
      *Context:* There are three exam slots available per day.
      Solutions must have a conflicts value of 0 or 1.

      *Tier 1:* Minimize Mandatory Reschedules
      The highest priority is to minimize the metrics associated with intense
      exam clusters that require a mandatory reschedule. Please minimize the
      following metrics in the order listed:
          quints (five exams in 24 hours)
          quads (four exams in 24 hours)
          four in five slots
          triple in 24h (no gaps)
          triple in same day (no gaps)

      *Tier 2:* Reduce Student Exam Fatigue
      Once Tier 1 objectives have been met, the primary tie-breaker is to
      minimize the total number of back-to-back exams. This objective is
      achieved by minimizing the sum of the following two metrics:
          evening/morning b2b
          other b2b

      *Tier 3:* Enhance Overall Schedule Quality
      As a final set of tie-breakers, prioritize schedules that improve the
      general student experience by optimizing these metrics in order of
      preference:
          Lower the avg_max (the average slot of a student's final exam).
          Minimize two in three slots.
          Minimize three in four slots.

Use these definitions (lower is better unless stated otherwise):
    - conflicts: 2 exams at the same time (requires makeup)
    - quints: 5 in a row (requires makeup)
    - quads: 4 in a row (requires makeup)
    - four in five slots: 4 within 5 (requires makeup)
    - triple in 24h (no gaps): 3 within 24h, consecutive (requires makeup)
    - triple in same day (no gaps): 3 in one day, consecutive (requires makeup)
    - three in four slots: 3 within 4 slots
    - evening/morning b2b: evening into next morning
    - other b2b: any other back-to-back pair
    - two in three slots: 2 within 3 slots
    - avg_max: average slot of each student's last exam (lower is earlier finish)

\end{verbatim}
\end{tcolorbox}
\captionof{figure}{An example prompt for the exam scheduling dataset.}
\label{fig:utility_prompt}

\subsection{Flights}
Similar to the exam scheduling prompt, in Figure \ref{fig:flight_prompt_hard}, the persona context is provided in the first sentence. The next paragraph states the user's preferences in natural language (``I am looking for a round-trip flight...''), and the attribute definitions follow below.

\begin{tcolorbox}[title=Flights Prompt,
                  colback=gray!5!white, colframe=gray!70!black,
                  sharp corners, boxrule=0.6pt, width=\textwidth,
                  breakable]
\begin{verbatim}
You are an expert travel scheduling agent that specializes in air fare.
    The data set is only for a one-way flight. In this case, it is the first leg
    of a round trip.

Objective: I am looking for round-trip flight options for two adults from Chicago
to New York City for the weekend of October 11, 2025. The priority is to maximize
our time in NYC while respecting a key scheduling constraint on Friday.

Primary Requirements (Must-Haves):
    - Route: Chicago (any airport) to New York City (JFK, LGA, or EWR).
    - Passengers: 2 adults.
    - Departure Window:
    - Must depart on either Friday, October 10, 2025, or Saturday,
    October 11, 2025.
    - If departing on Friday, the flight must be after 12:30 PM.
    - Return Date: Sunday, October 12, 2025.
    - Budget: The total price per ticket must not exceed $400.

Secondary Preferences (Tie-Breakers):
    - These should be used to select the best option among flights that meet the
    above requirements.
    - Departure Time (Friday): A departure after 3:30 PM is strongly preferred.
    The ideal arrival would be early enough to get a full night's sleep or to
    catch an 11:00 PM train from the airport.
    - Departure Time (Saturday): If a Saturday departure is chosen, it should be
    the earliest possible flight to maximize time in the city.
    - Flight Type: Direct (non-stop) flights are highly preferred.
    - Cost: Among flights that meet all criteria, the cheapest option is best.
    - Destination Airport: There is a slight preference against Newark (EWR) due
    to longer ground transportation, but it is an acceptable option.
    - Airline: No airline preference.

    Use these definitions:
    • name: name of airline operating the flight
    • origin: origin airport
    • destination: destination airport
    • departure time: time of departure from origin airport
    • arrival time: time of arrival at destination airport
    • duration: how long the flight is
    • stops: number of layover stops
    • price: cost of the flight
    • dis_from_origin: distance of origin airport from where the customer prefers
    (lower is better)
    • dis_from_dest: distance of arrival airport from where the customer prefers
    (lower is better)
    • departure_seconds: time of departure since a fixed date in seconds
    • arrival_seconds: time of arrival since a fixed date in seconds
    • duration_min: duration of total flight in minutes (lower is better)
\end{verbatim}
\end{tcolorbox}
\captionof{figure}{An example prompt for a flight itinerary preference, which utilizes the natural language description of the user's priorities over objectives, but includes strong constraints.}
\label{fig:flight_prompt_hard}

\subsection{Headphones}
Following the format of the other prompts, the first line in Figure \ref{fig:soft_headphones_prompt} details the persona context (``You are an audio equipment reviewer..."). The next section details the user's preferences in natural language, and the attribute definitions follow below. These are all ``soft" preferences because they do not include any strict constraints.
Figure \ref{fig:strict_headphones_prompt} has the same persona context, but the user's preferences are organized by must-haves and nice-to-haves. This is the main difference between the soft and strict headphones prompt. For example, in the strict headphones prompt, the user requires that the pair of chosen headphones has a microphone, but is flexible on price, with cheaper options being a nice-to-have but not a strict requirement. On the other hand, in the soft headphones prompt, the user simply uses phrases like ``I prefer" rather than ``I must have".

\begin{tcolorbox}[title=Soft Headphones Prompt,
                  colback=gray!5!white, colframe=gray!70!black,
                  sharp corners, boxrule=0.6pt, width=\textwidth,
                  breakable]
\begin{verbatim}
You are an audio equipment reviewer choosing the best headphones.

I am a college student who is looking for recommendations for new headphones. I
prefer over-ear headphones rather than in-ear or on-ear, since they feel
more comfortable for long hours and usually deliver better sound. I also prefer
wireless headphones because I don’t want to deal with cables. I would like
the headphones to have active noise cancellation, since I often study in
places where there is background noise. A microphone is important to me because
I sometimes use my headphones for calls. For battery life, I’d like
headphones that can last a long time on a single charge.
I usually don’t mind recharging overnight, but I don’t want headphones that
run out of power quickly in the middle of the day.
In terms of build, I actually prefer headphones that feel a bit heavier, since
lightweight headphones sometimes feel cheap or less durable to me. When I look
at reviews, I prefer headphones with both a high rating and a large number of
reviews, since that gives me more confidence in the product. I would prefer
headphones which have at least 5,000 reviews and a rating of at least 4.4.
To clarify the relative importance of both rating and number of reviews, I would
rather pick headphones with a 4.8 rating and 20,000 reviews than a 4.9 rating
with 1000 reviews. Other metrics such as driver size, water resistance, and
Bluetooth version do not matter at all to me. I’m not price-sensitive at all,
so I’m willing to pay more if the headphones meet my preferences well.

    Use these definitions:
    - product_name: name of the product (non-metric, weight always 0).
    - brand: manufacturer reputation (non-metric, weight always 0).
    - price: cost measured in dollars.
    - type: headphone design.
    - connectivity: wired vs wireless.
    - noise_cancellation: active vs passive.
    - battery_life: measured in hours.
    - bluetooth_version: higher versions are newer.
    - driver_size: measured in millimeters.
    - weight: measured in ounces.
    - water_resistance: measured subjectively and also through the ipXX rating
        system, which measures a device's protection against solids (first digit)
        and liquids (second digit), with higher numbers indicating stronger
        resistance.
    - microphone: presence/quality of mic
    - review_rating: average customer rating
    - review_count: number of reviews
    - description: qualitative marketing text (non-metric, weight always 0)

\end{verbatim}
\end{tcolorbox}
\captionof{figure}{Soft prompt for headphone preferences.}
\label{fig:soft_headphones_prompt}

\begin{tcolorbox}[title=Strict Headphones Prompt,
                  colback=gray!5!white, colframe=gray!70!black,
                  sharp corners, boxrule=0.6pt, width=\textwidth,
                  breakable]
\begin{verbatim}
You are an audio equipment reviewer choosing the best headphones.

I am seeking recommendations for premium wireless over-ear headphones designed
for both long study sessions and high-quality music listening. The ideal options
should balance advanced features, comfort, and practicality to support a student
lifestyle.
Primary Requirements (Must-Haves):
HARD CONSTRAINTS:
  - type MUST BE 6 (Over-Ear)
  - connectivity MUST BE 2 (Wireless)
  - noise_cancellation MUST BE 1 (Active)
  - microphone MUST BE 2 (Yes)
  - battery_life MUST BE >= 30 hours
Type: Wireless, over-ear headphones (not earbuds).
Core Features: Active Noise Cancellation (ANC) and a built-in microphone.
Battery Life: Minimum 30 hours of playback per charge to ensure multi-day use
without nightly recharging.
Comfort & Weight: A balanced design—substantial enough to feel durable and
premium, yet light enough for several hours of wear without causing head
or neck strain.
Brand: At least one recommendation must feature a Sennheiser model.
User Reviews: Preference for models with both a high average rating and
a large number of reviews, signaling broad user satisfaction.

Secondary Criteria (Nice to Have):
Product Information: Descriptions should highlight either superior sound
quality or exceptional long-wear comfort.
Price: When two or more options meet all requirements equally, the more
affordable option is preferred.
Battery Life: Minimum 30 hours of playback per charge to ensure multi-day
use without nightly recharging.

    Use these definitions:
    - product_name: name of the product (non-metric, weight always 0).
    - brand: manufacturer reputation (non-metric, weight always 0).
    - price: cost measured in dollars.
    - type: headphone design.
    - connectivity: wired vs wireless.
    - noise_cancellation: active vs passive.
    - battery_life: measured in hours.
    - bluetooth_version: higher versions are newer.
    - driver_size: measured in millimeters.
    - weight: measured in ounces.
    - water_resistance: measured subjectively and also through the ipXX rating
        system, which measures a device's protection against solids (first digit)
        and liquids (second digit), with higher numbers indicating stronger
        resistance.
    - microphone: presence/quality of mic
    - review_rating: average customer rating
    - review_count: number of reviews
    - description: qualitative marketing text (non-metric, weight always 0)

\end{verbatim}
\end{tcolorbox}
\captionof{figure}{Strict prompt for headphone preferences.}
\label{fig:strict_headphones_prompt}

\section{Value of the Preference Utterance}

\begin{figure}[H]
    \centering
    \begin{subfigure}{0.49\linewidth}
        \centering
        \includegraphics[width=\linewidth]{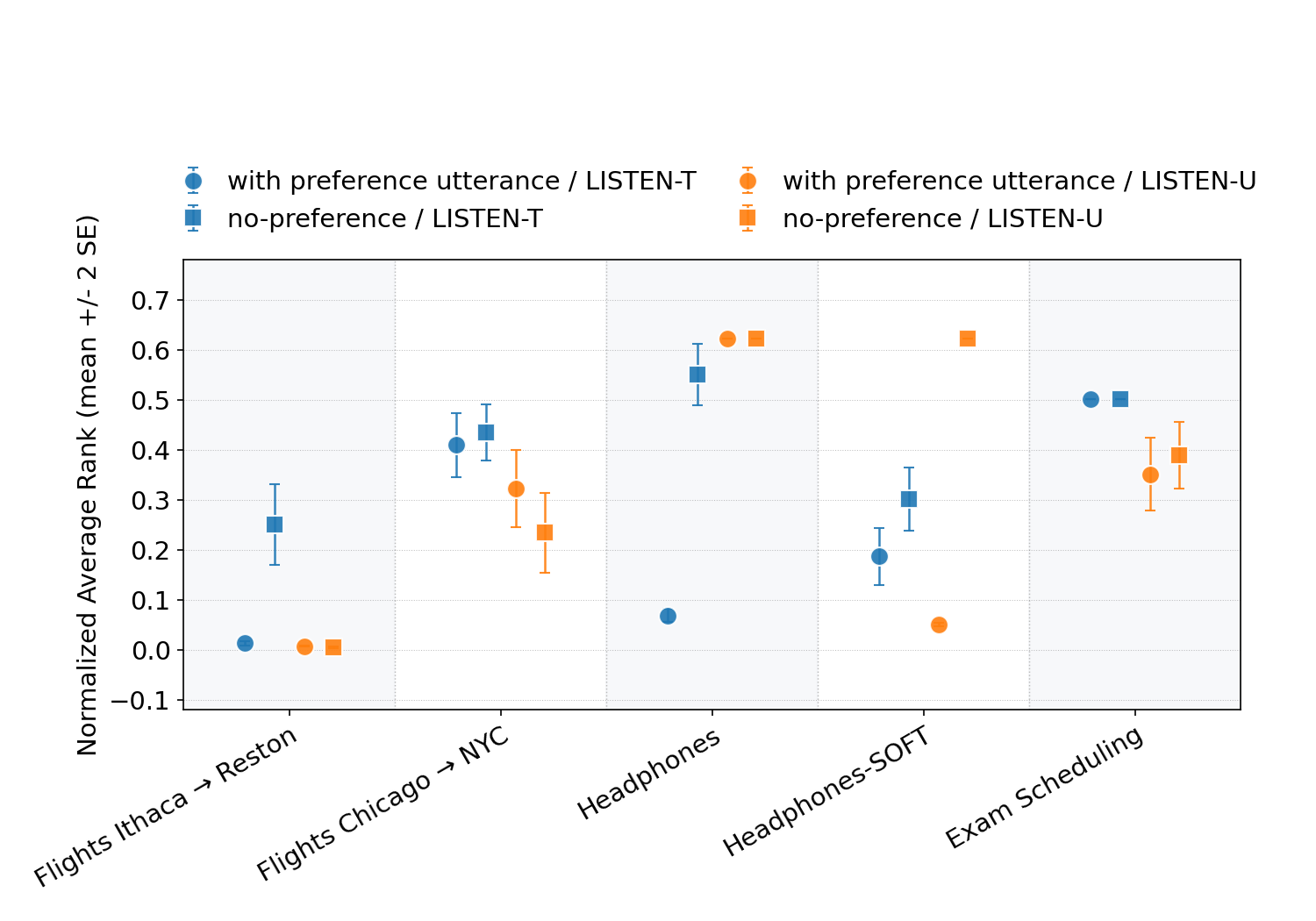}
        \caption{LLaMA.}
        \label{fig:prompt-format-base-llama}
    \end{subfigure}
    \hfill
    \begin{subfigure}{0.49\linewidth}
        \centering
        \includegraphics[width=\linewidth]{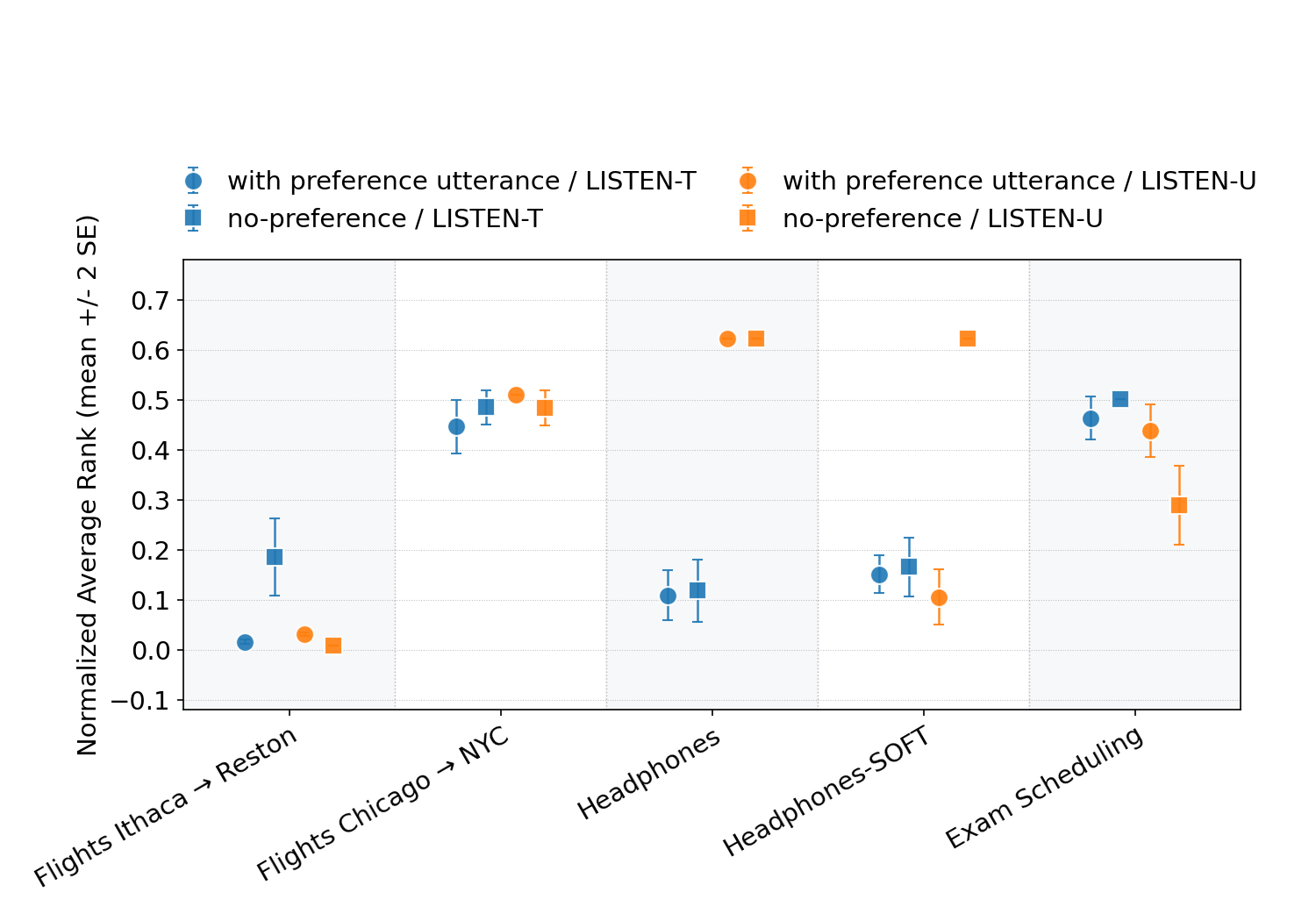}
        \caption{Gemini.}
        \label{fig:prompt-format-base-gemini}
    \end{subfigure}
    \caption{Performance of LISTEN-U and LISTEN-T with and without the preference utterance. Each bar averages $n=40$ runs.}
    \label{fig:prompt-format-base}
\end{figure}

To isolate the contribution of the user's natural language preference utterance ($U$), we conducted an ablation study comparing two prompt variants: the full, preference-guided prompt used throughout the paper, and a \textit{base prompt} containing only the persona and attribute definitions (i.e., with $U$ removed). The results are shown in Figure~\ref{fig:prompt-format-base}.

The value of the preference utterance is model-dependent. For LLaMA, adding $U$ improves LISTEN-T across all four datasets and generally improves LISTEN-U as well, although the size of the gain varies by task. For Gemini, the effect is less consistent: in most settings, the full and base prompts have overlapping error bars, so the figure does not show a clear separation between them. The clearest gains for Gemini appear in LISTEN-T on Flights Ithaca $\rightarrow$ Reston and LISTEN-U on Headphones-Soft. This suggests that the benefit of $U$ depends not only on the selection algorithm, but also on the model's ability to use a natural-language preference description reliably. One possible explanation is that Gemini 2.5 Flash-Lite is a relatively small, efficiency-oriented model, and may extract less stable benefit from the additional preference text. A broader comparison across model sizes would be needed before drawing stronger conclusions.

\section{Prompt Order Robustness}
\label{sec:prompt-order}
\begin{figure}[H]
    \centering
    \begin{subfigure}{\linewidth}
        \centering
        \includegraphics[width=\linewidth]{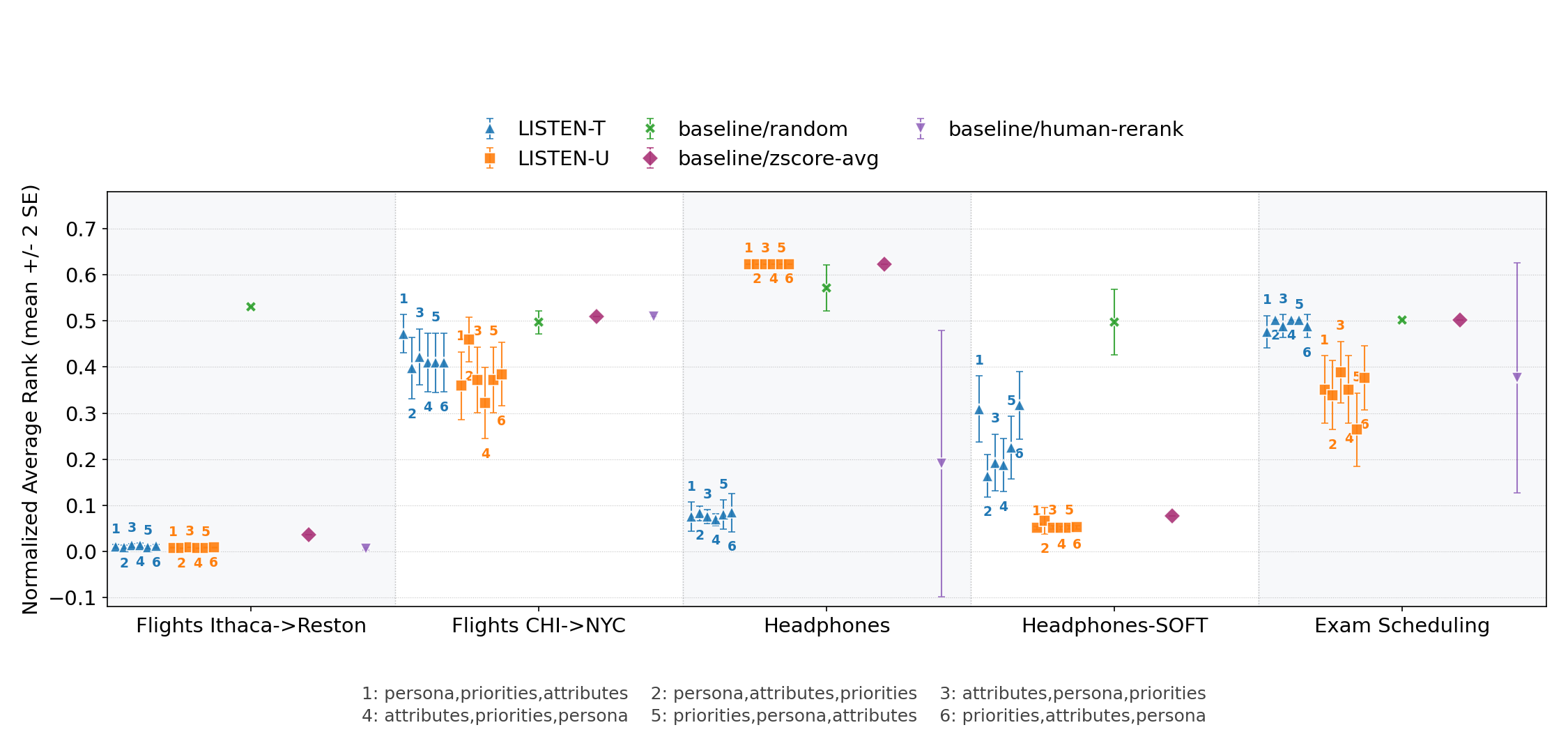}
        \caption{LLaMA.}
        \label{fig:prompt-format-llama}
    \end{subfigure}

    \vspace{0.5em}

    \begin{subfigure}{\linewidth}
        \centering
        \includegraphics[width=\linewidth]{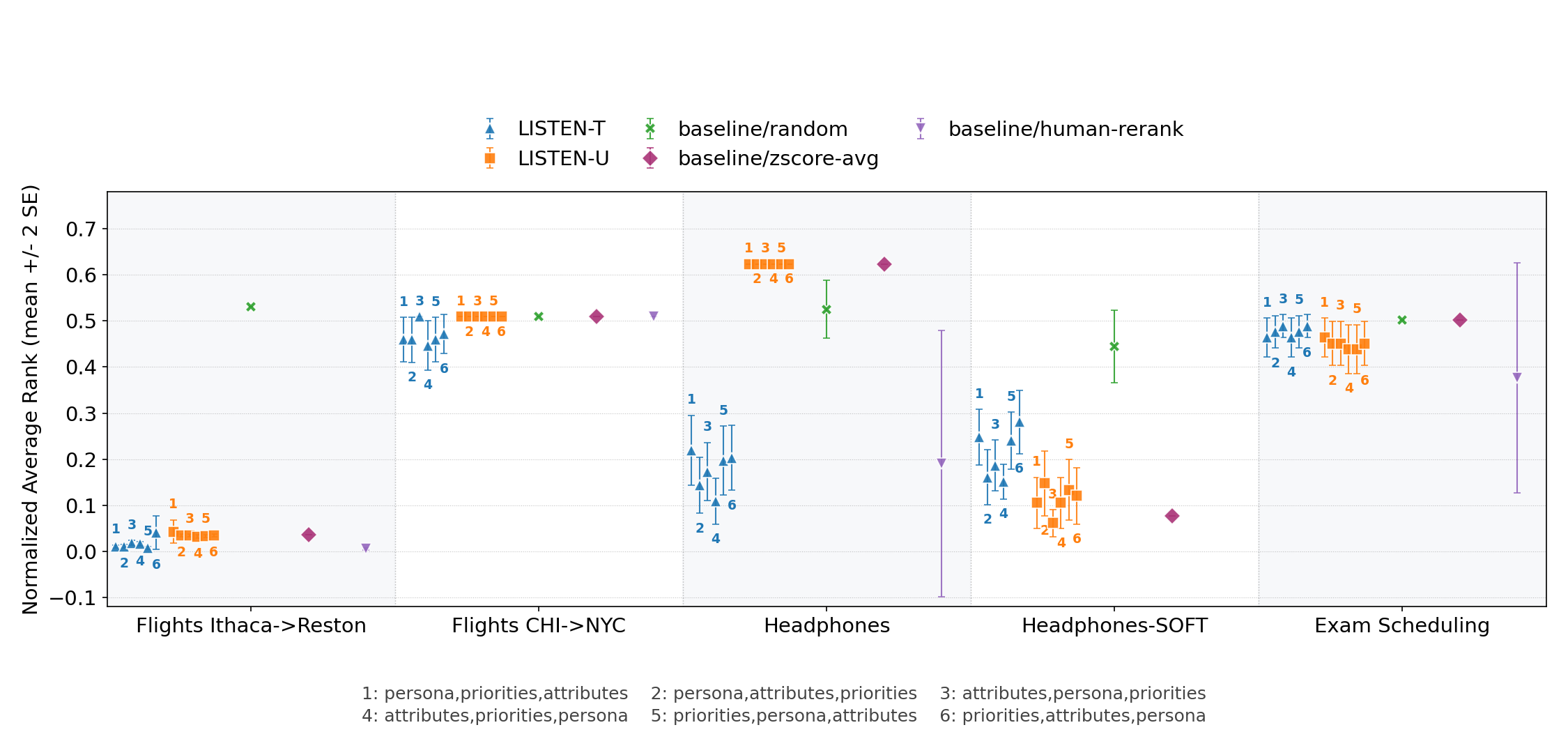}
        \caption{Gemini.}
        \label{fig:prompt-format-gemini}
    \end{subfigure}
    \caption{Prompt order ablation. Normalized Average Rank (lower is better) is shown as mean $\pm$ 2 SE under all six permutations of \emph{Persona Context}, \emph{Attribute Definitions}, and \emph{User Priorities}. Points average $n{=}40$ runs for each algorithm except baseline-zscore ($n{=}1$, deterministic) and human rerank ($n{=}4$).}
    \label{fig:prompt-format}
\end{figure}

To assess sensitivity to prompt layout, we permute three of the five prompt components defined in Section~3.1: \emph{Persona Context}, \emph{Attribute Definitions}, and \emph{User Priorities}. Figure~\ref{fig:prompt-format} reports Normalized Average Rank for all six orderings, labeled 1--6 in the legend. The ranking of methods is preserved across these permutations.

\section{Additional Results by Batch Size}
\label{app:batch}

\begin{figure}[!ht]
    \centering
    \begin{subfigure}{0.49\linewidth}
        \centering
        \includegraphics[width=\linewidth]{figs/TournamentExperiment__exam__groq__nar__batch_size.png}
        \caption{LLaMA, Exam Scheduling}
        \label{fig:exam-llama}
    \end{subfigure}
    \hfill
    \begin{subfigure}{0.49\linewidth}
        \centering
        \includegraphics[width=\linewidth]{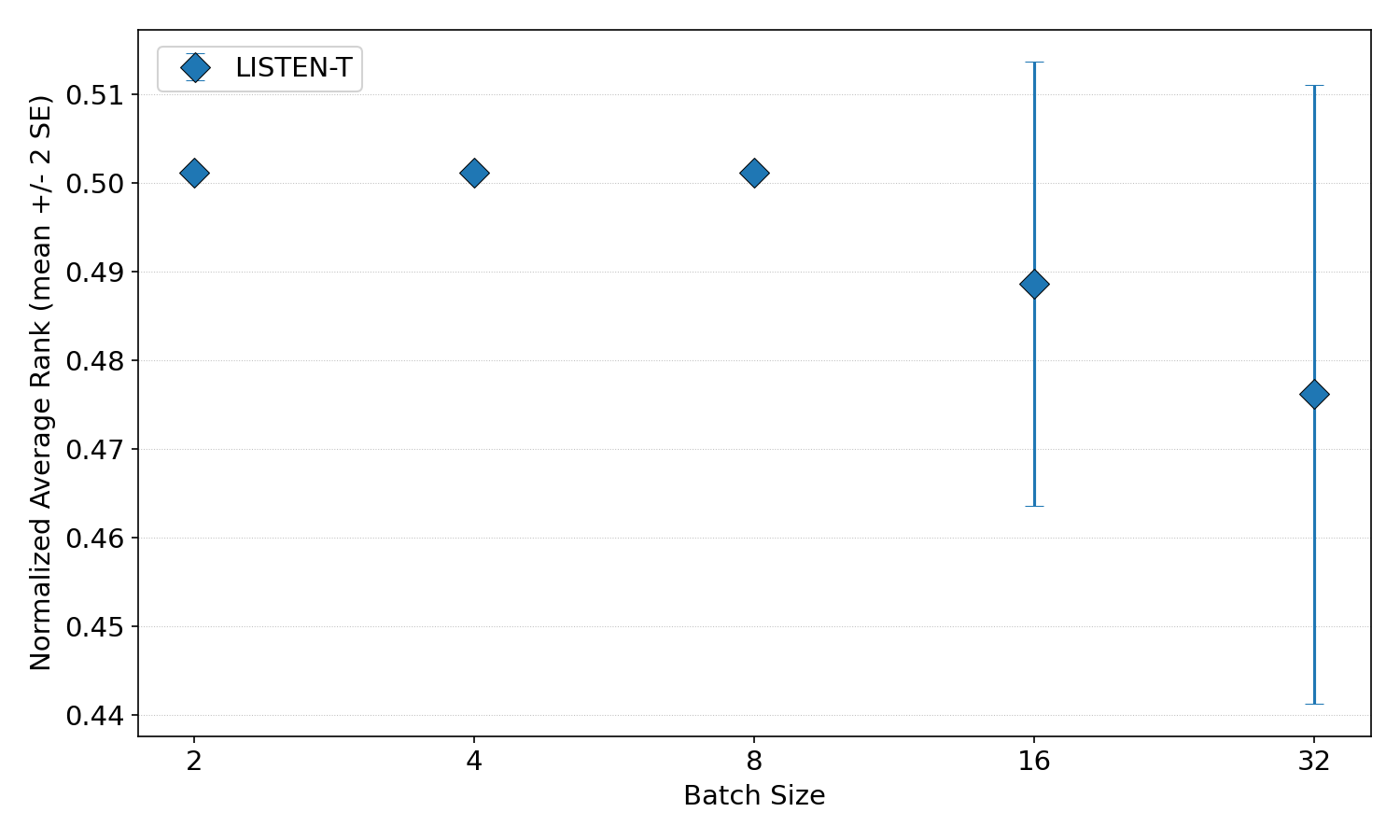}
        \caption{Gemini, Exam Scheduling}
        \label{fig:exam-gemini}
    \end{subfigure}

    \vspace{0.5em}

    \begin{subfigure}{0.49\linewidth}
        \centering
        \includegraphics[width=\linewidth]{figs/TournamentExperiment__flights_chi_nyc__groq__nar__batch_size.png}
        \caption{LLaMA, Flights CHI $\rightarrow$ NYC}
        \label{fig:flight00-llama}
    \end{subfigure}
    \hfill
    \begin{subfigure}{0.49\linewidth}
        \centering
        \includegraphics[width=\linewidth]{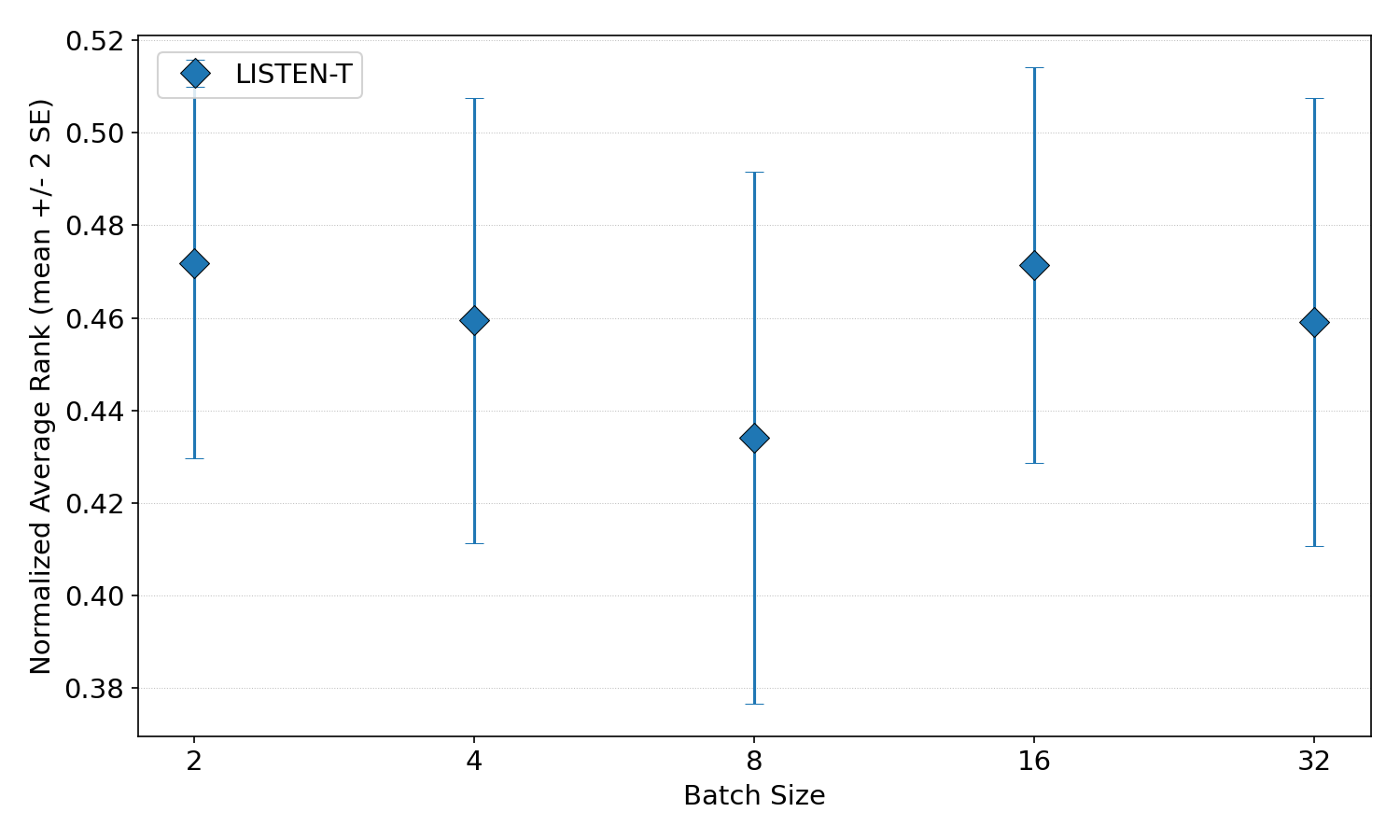}
        \caption{Gemini, Flights CHI $\rightarrow$ NYC}
        \label{fig:flight00-gemini}
    \end{subfigure}

    \vspace{0.5em}

    \begin{subfigure}{0.49\linewidth}
        \centering
        \includegraphics[width=\linewidth]{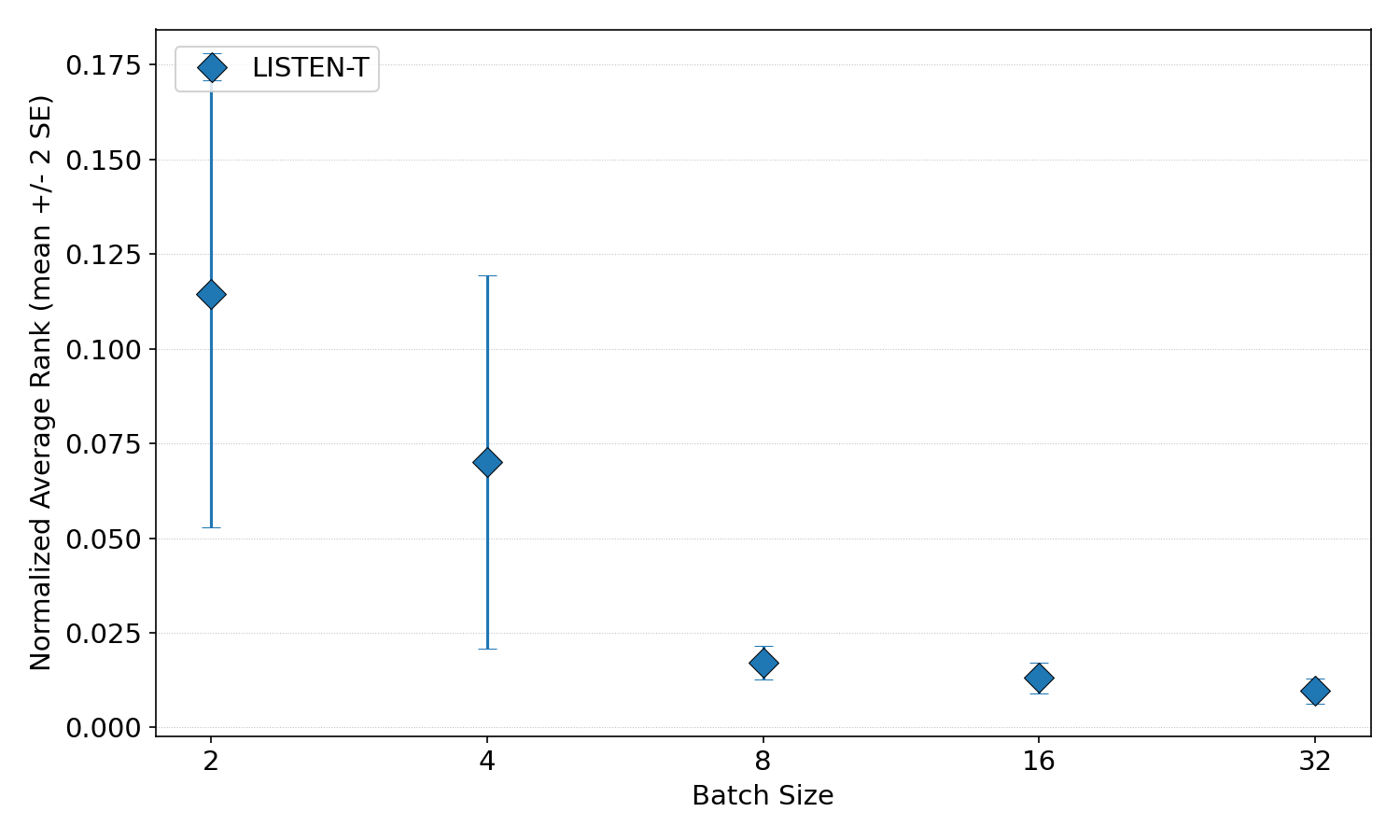}
        \caption{LLaMA, Flights Ithaca $\rightarrow$ Reston}
        \label{fig:ith-llama}
    \end{subfigure}
    \hfill
    \begin{subfigure}{0.49\linewidth}
        \centering
        \includegraphics[width=\linewidth]{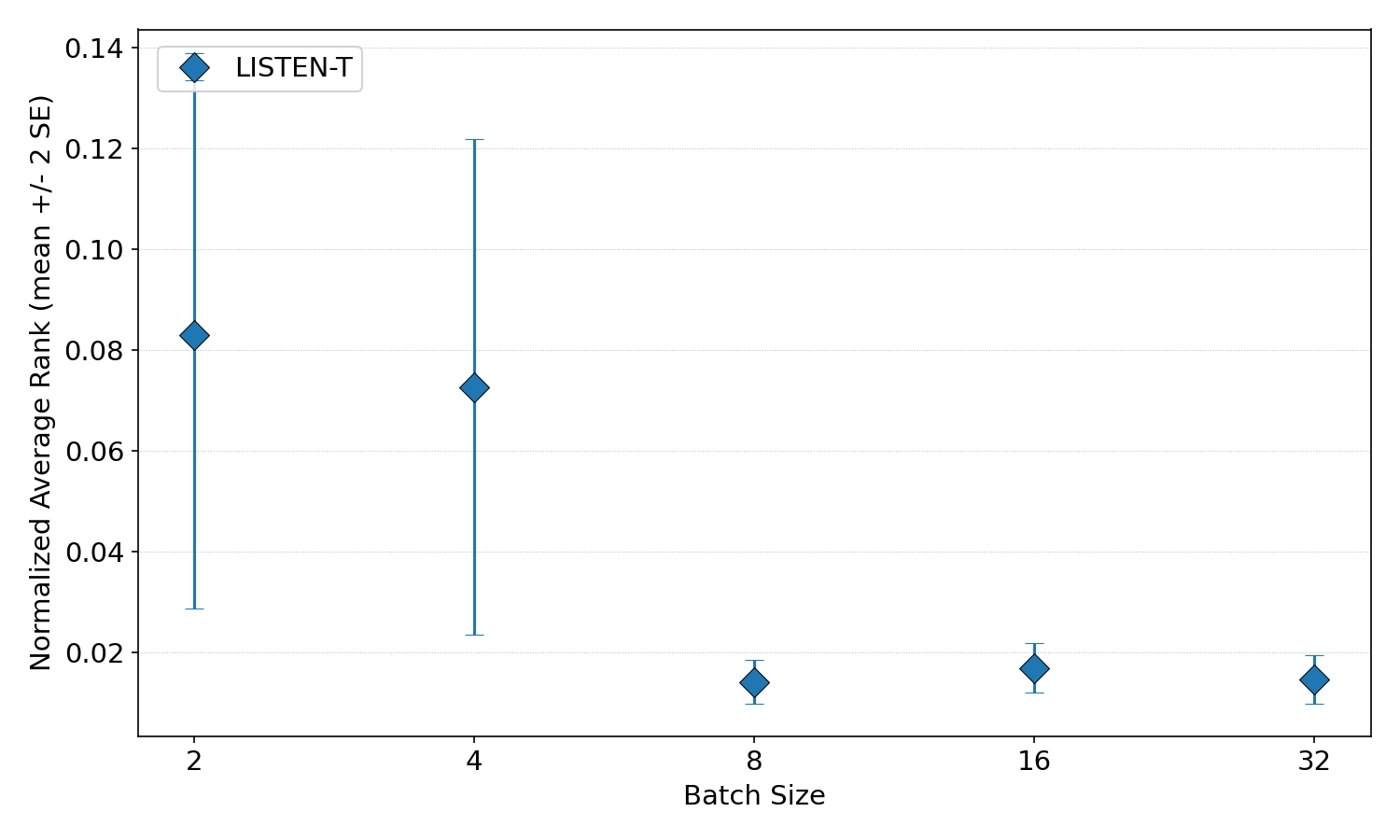}
        \caption{Gemini, Flights Ithaca $\rightarrow$ Reston}
        \label{fig:ith-gemini}
    \end{subfigure}

    \vspace{0.5em}

    \begin{subfigure}{0.49\linewidth}
        \centering
        \includegraphics[width=\linewidth]{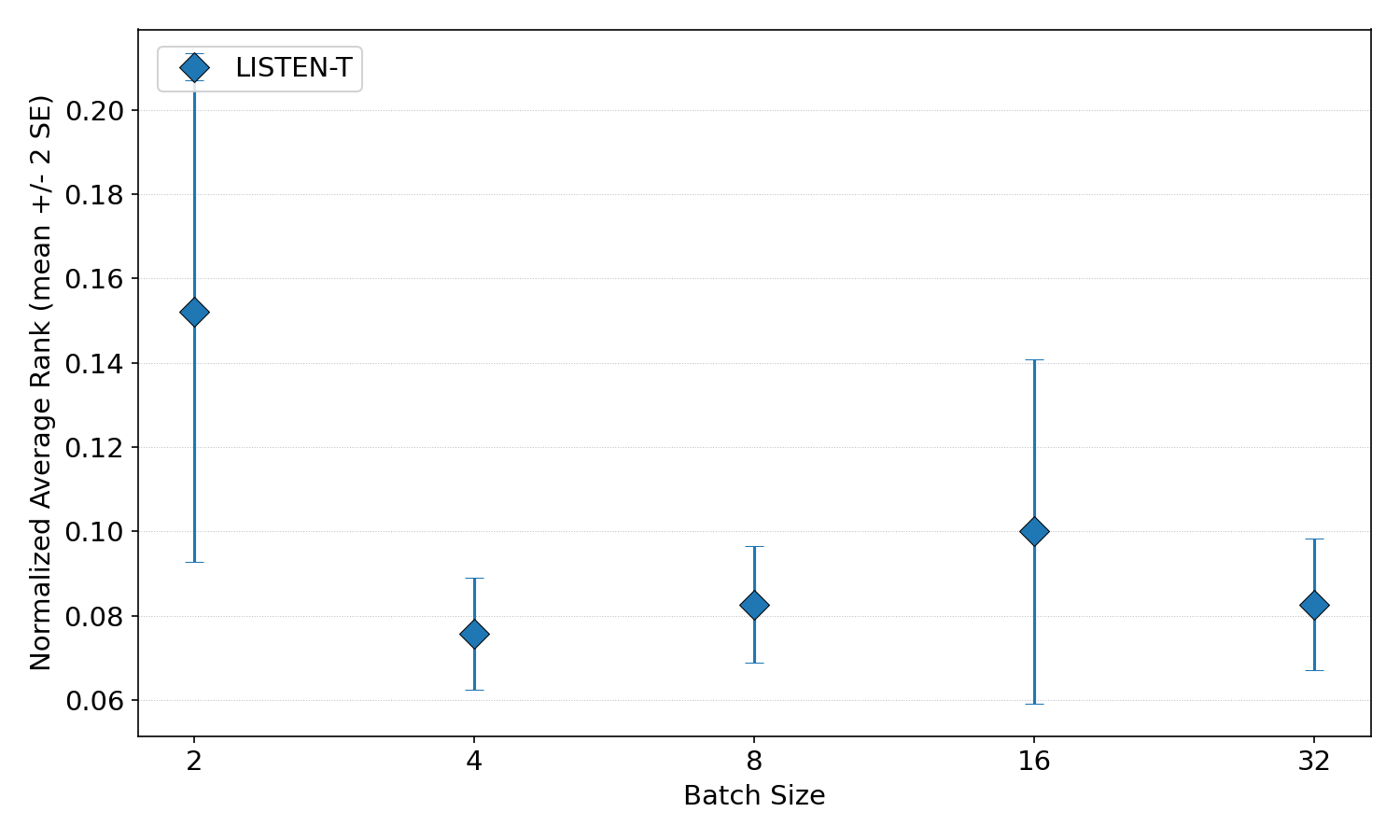}
        \caption{LLaMA, Headphones}
        \label{fig:headphones-llama}
    \end{subfigure}
    \hfill
    \begin{subfigure}{0.49\linewidth}
        \centering
        \includegraphics[width=\linewidth]{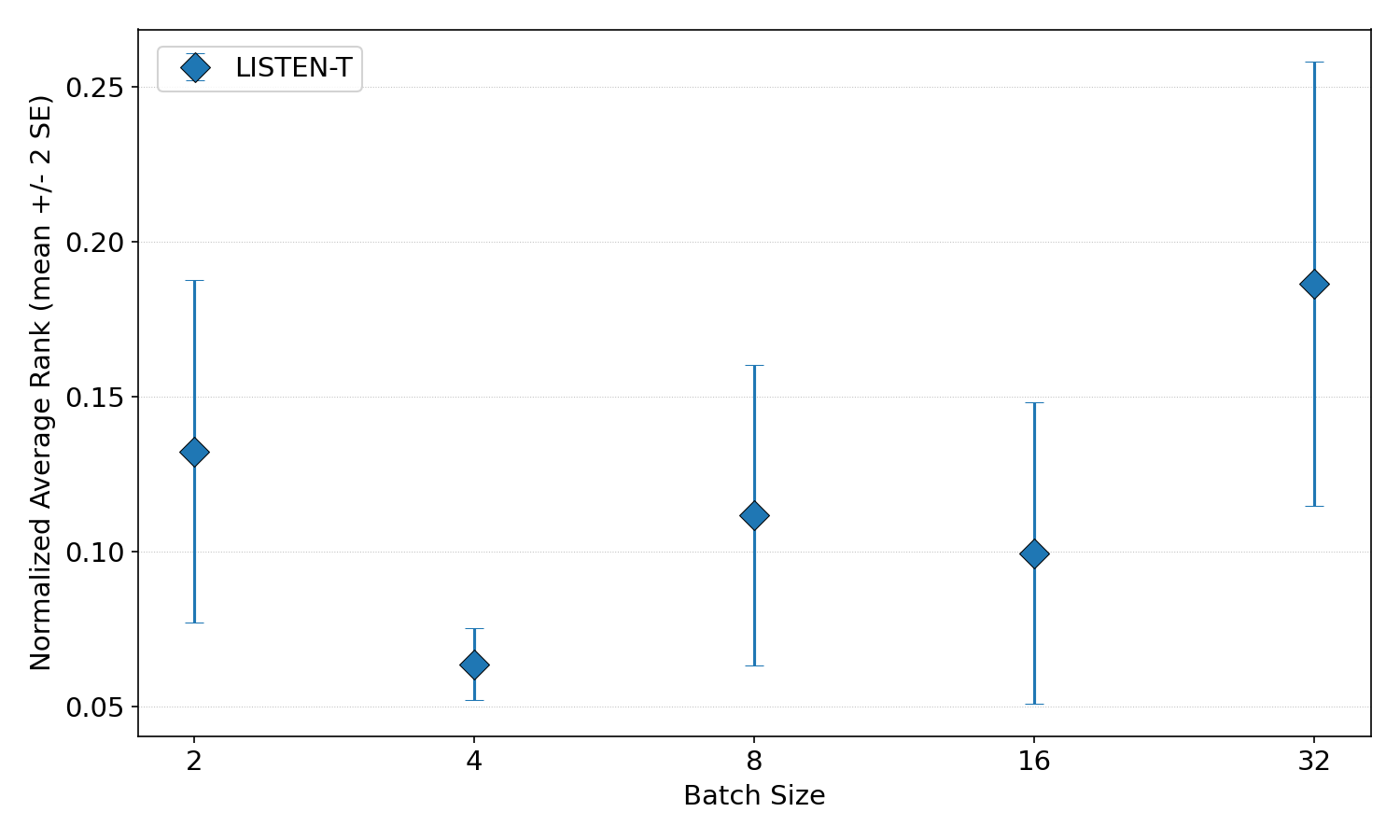}
        \caption{Gemini, Headphones}
        \label{fig:headphones-gemini}
    \end{subfigure}

    \caption{LISTEN-T batch size effect across tasks for LLaMA vs.\ Gemini. Each row shows the same problem instance, with LLaMA (left) and Gemini (right). Each point averages $n=40$ runs per batch size.}
    \label{fig:bs-all}
\end{figure}

Figure~\ref{fig:bs-all} reports LISTEN-T's Normalized Average Rank as a function of batch size $B$ for each (LLM, dataset) pair. For most pairs the ranking is essentially flat in $B$: changing the batch size does not meaningfully change the rank of the LISTEN-T champion. The one clear exception is \textbf{Flights Ithaca $\rightarrow$ Reston}, where both LLaMA and Gemini improve as $B$ grows. A plausible explanation is concordance: Flights Ithaca $\rightarrow$ Reston has the highest concordance score of the four datasets (Table~\ref{tab:difficulty}), meaning its human-preferred items are well-aligned with simple scoring rules the LLM can apply consistently. In that regime, widening LISTEN-T's per-round batch both raises the chance that a top-ranked item is present and gives the LLM more chances to correctly identify it as the batch champion. On the lower-concordance datasets, preferences hinge on hard constraints or non-linear trade-offs that the LLM struggles to apply reliably, so simply showing it more candidates per round does not translate into rank improvements.
Dataset size is a secondary confound worth flagging: Headphones' small catalog ($N{=}77$) means batch-32 already covers ${\sim}40\%$ of the items per round, leaving less room for $B$ to grow than on Flights Ithaca $\rightarrow$ Reston ($N{=}216$, ${\sim}15\%$ coverage). However, the two largest datasets, Exam Scheduling ($N{=}4938$) and Flights CHI $\rightarrow$ NYC ($N{=}903$), also show flat curves despite having far more headroom, which suggests concordance is the dominant driver rather than catalog size.

\end{document}